\documentclass{article}

\PassOptionsToPackage{sort, compress, numbers}{natbib}
\usepackage[preprint]{nips_2018}

\usepackage[utf8]{inputenc} 
\usepackage[T1]{fontenc}    
\usepackage{hyperref}       
\usepackage{url}            
\usepackage{booktabs}       
\usepackage{amsfonts}       
\usepackage{nicefrac}       
\usepackage{microtype}      %
\usepackage{enumitem}
\usepackage{algorithm,algpseudocode}
\usepackage{amsmath,amsthm,graphicx,tabularx,epstopdf,url}

\makeatletter
\g@addto@macro\normalsize{%
  \setlength\abovedisplayskip{2pt}
  \setlength\belowdisplayskip{2pt}
  \setlength\abovedisplayshortskip{1pt}
  \setlength\belowdisplayshortskip{1pt}
}
\makeatother

\usepackage[skip=2mm]{caption}
\usepackage{xcolor}
\hypersetup{
    colorlinks,
    linkcolor={red!80!black},
    citecolor={blue!50!black},
    urlcolor={blue!90!black}
}

\def\bw{\mathbf{w}}

\def\bb{\mathbf{b}}

\def\bx{\mathbf{x}}
\def\by{\mathbf{y}}

\def\bphi{\boldsymbol{\phi}}
\def\bPhi{\boldsymbol{\Phi}}

\def\rone{\overrightarrow{\mathbf{1}}}

\def\indSet{\mathcal{I}}

\def\bA{\mathbf{A}}
\def\bB{\mathbf{B}}
\def\bC{\mathbf{C}}
\def\bM{\mathbf{M}}
\def\bU{\mathbf{U}}
\def\bW{\mathbf{W}}
\def\bX{\mathbf{X}}
\def\bY{\mathbf{Y}}

\def\b0{\mathbf{0}}

\def\bK{\mathbf{K}}

\def\bSigma{\boldsymbol{\Sigma}}

\def\lnorm{\left\|}
\def\rnorm{\right\|}
\def\lp{\left(}
\def\rp{\right)}

\DeclareMathOperator{\concat}{concat}

\DeclareMathOperator{\blkdiag}{blkdiag}

\theoremstyle{plain}
\newtheorem{prop}{Proposition}

\usepackage{color}

\newcolumntype{Y}{>{\centering\arraybackslash}X}

\title{Supervising Nystr\"{o}m Methods via Negative Margin Support Vector Selection}

\author{
  Mert Al\textsuperscript{1},\quad Thee Chanyaswad\textsuperscript{2},\quad Sun-Yuan Kung\textsuperscript{3}
  \\
  Department of Electrical Engineering\\
  Princeton University\\
  Princeton, New Jersey, USA \\
  \texttt{\textsuperscript{1}merta@princeton.edu, \textsuperscript{2}tc7@princeton.edu, \textsuperscript{3}kung@princeton.edu} \\
}

\begin{document}
\maketitle

\begin{abstract}
 The Nystr\"om methods have been popular techniques for scalable kernel based learning. They approximate explicit, low-dimensional feature mappings for kernel functions from the pairwise comparisons with the training data. 
However, Nystr\"{o}m methods are generally applied without the supervision provided by the training labels in the classification/regression problems. This leads to pairwise comparisons with randomly chosen training samples in the model.
Conversely, this work studies a supervised Nystr\"{o}m method that chooses the critical subsets of samples for the success of the Machine Learning model. Particularly, we select the \emph{Nystr\"{o}m support vectors} via the \emph{negative margin} criterion, and create explicit feature maps that are more suitable for the classification task on the data. Experimental results on six datasets show that, without increasing the complexity over unsupervised techniques, our method can significantly improve the classification performance achieved via kernel approximation methods 
and reduce the number of features needed to reach or exceed the performance of the full-dimensional kernel machines.
\end{abstract}

\section{Introduction}
Kernel methods have been successful in various applications, e.g. \cite{scholkopf2004kernel,joachims2002learning,3schuller2004speech,4osuna1997training,5guyon2002gene}.
Their main innovation is the mapping of the data onto a high-dimensional feature space, without having to compute the expansions explicitly \cite{vapnik2013nature,kung2014kernel}. This is achieved via the kernel trick, which only requires a Gram (kernel) matrix to be computed in the original feature space. Given $N$ training samples, the kernel matrix is $N \times N$. Hence, although this may be advantageous for small-scaled applications,
for large-scaled learning \textendash{} where $N$ can be massive \textendash{} the size of the kernel matrix quickly becomes an obstacle. Previous work has addressed this challenge primarily via kernel matrix approximation \cite{girolami2002orthogonal,drineas2005nystrom,rahimi2008random}. These methods lead to explicit, low-dimensional, approximate representations of the implicit, high-dimensional mappings for the data. Problems such as classification and regression can then be solved via the primal domain algorithms working in the approximate feature space for the kernel, as opposed to the dual domain algorithms usually used in the kernel machines.

Kernel approximation methods can be grouped under two categories; data dependent \cite{drineas2005nystrom,zhang2008improved,kumar2009ensemble,
li2010making,si2014memory,li2015large} and data independent \cite{rahimi2008random,rahimi2009weighted,le2013fastfood,yang2015carte}. In this work, we focus on the data dependent approach. Notably, the data dependent (Nystr\"om) methods conceptually perform Kernel Principal Component Analysis (KPCA) with random subsets of training data to create the explicit feature mappings. As a result, they lead to decision makers, which are functions of pairwise comparisons with the training data. Yet, unlike the margin maximizing Support Vector Machines (SVMs), the so called \emph{support vectors} of these models are chosen independently from the pattern recognition task.

One of the key advantages of SVM models is that they place greater emphasis on the more important samples, i.e., the support vectors \cite{tao2006asymmetric,kung2008feature,vapnik2013nature}.  Inspired by this, we propose the use of a supervised sample selection method to enhance the Nystr\"{o}m methods, as opposed to the traditional unsupervised Nystr\"{o}m variants. More specifically, we employ a two stage procedure. In the first stage, an approximate kernel classifier is trained using standard Nystr\"{o}m techniques. In the second stage, support vectors are chosen based on the classifier from the first stage. These are then used to extract features that are more suitable for the classification task. To the best of our knowledge, this is the first work that chooses the subsets of samples used by Nystr\"{o}m methods in a supervised manner.

Due to the relation of this objective to sample importance weighting for 
multiple Machine Learning models, we propose the \emph{negative margin} criterion for the selection of the support vectors. Specifically, the negative margin criterion measures how far on the wrong side of the classification boundary a sample lies. By exploiting the ability of Nystr\"{o}m methods to approximate this quantity, our two stage procedure allows them to restrict the solution to a more optimal subspace, without increasing the complexity. The experimental results on six datasets demonstrate that, not only can support vector selection improve the classification performance of kernel approximation methods, but it can help exceed the performances of full-dimensional Kernel Ridge Regression and SVM as well.

\subsection{Related Work}
Many data independent approximations of kernel based features have been proposed. Rahimi and Recht introduced random features to approximate shift invariant kernels \cite{rahimi2008random,rahimi2009weighted}. The most well-known of such techniques is the \emph{Random Fourier Features}. 
These methods were later extended for improved complexity and versatility \cite{le2013fastfood,yang2015carte}. Random features have desirable generalization properties \cite{rudi2017generalization}, even though data dependent approximations were shown to exploit the structure in the data better, both theoretically and empirically \cite{yang2012nystrom,rudi2017generalization}.

The Standard Nystr\"{o}m algorithm can be viewed as the application of KPCA to a small, randomly chosen subset of training samples \cite{girolami2002orthogonal}. Various works have altered this method to achieve better approximations with less memory/computation. Zhang \textit{et al.} use k-means centroids to perform KPCA, instead of a random subset of the data \cite{zhang2008improved}. Kumar \textit{et al.} combine multiple smaller scale KPCAs \cite{kumar2009ensemble}. Li \textit{et al.} utilize randomized SVD to speed up KPCA for the Nystr\"{o}m algorithms \cite{li2015large}. Additionally, non-uniform sampling schemes have been explored to improve the performance of Nystr\"{o}m \cite{drineas2005nystrom,drineas2012fast,gittens2016revisiting}. Though, these require at least one pass over the whole kernel matrix, resulting in $o(N^2)$ complexity. For linear and RBF kernels, uniform sampling has been shown to work well \cite{kumar2012sampling} and results in no additional complexity.

\emph{The support vector selection scheme we propose can be applied together with many Nystr\"{o}m variants,} such as those proposed in \cite{girolami2002orthogonal,kumar2009ensemble,li2015large}. The main difference of our approach from the previous work is that the subsets of samples used by Nystr\"{o}m methods are selected in a \emph{supervised manner}.

\section{Preliminaries}
 Our method consists of two stages, both of which utilize variants of the Nystr\"{o}m approximations. Therefore, we briefly discuss the variants we used here. More details about the variants other than Standard Nystr\"om can be found in Appendix \ref{app:variants}.
 
 
 \textit{Notation:} We denote by $\bK$ the full, $N \times N$ kernel matrix and by $\bPhi$ the full, $N$-columned data matrix in the kernel induced feature space. $\widetilde{\bK}$ and  $\widetilde{\bPhi}$ denote the approximations of the kernel and data matrices, respectively. Similarly, $K(\cdot,\cdot)$ and $\widetilde{K}(\cdot,\cdot)$ denote the kernel function and its approximation. $\mathcal{I}_n \subset \{1,\ldots,N\}$ denotes a subset of $n<N$ selected indices. We use MATLAB notation to refer to row and column subsets of a matrix, namely, $\bM(\indSet_n,:)$ and $\bM(:,\indSet_n)$ denote subsets of $n$ rows and columns of $\bM$, respectively, and $\bM(\indSet_n,\indSet_n)$ denotes an $n \times n$ dimensional submatrix of $\bM$. For a matrix $\bM$, we denote its best rank-$k$ approximation by $\bM_k$ and its Moore-Penrose inverse by $\bM^{+}$. $\lnorm \cdot \rnorm_2$ denotes the  the spectral norm and $\lnorm \cdot \rnorm_F$ denotes the Frobenius norm for matrices.
 
 \textbf{Standard Nystr\"{o}m} \cite{girolami2002orthogonal} algorithm projects the data into a kernel feature subspace spanned by $n \ll N$ samples. It produces a rank-$k$ approximation of the kernel matrix given by $\widetilde{\bK}=\bC\bB_k^+\bC^\top$, where $\bC=\bK(:,\indSet_n)$ and $\bB=\bK(\indSet_n,\indSet_n)$. This is the same as applying the non-centered KPCA feature mapping to the training data; $\widetilde{\bPhi}=\bSigma^{-\nicefrac{1}{2}}\bU^\top\bC^\top$, where $\bB_k=\bU\bSigma\bU^\top$ is the compact SVD of $\bB_k$.
 
Owing to the Representer Theorem \cite{wahba1990spline,scholkopf2001generalized}, applying the dual formulations of certain machine learning models using $\widetilde{\bK}$ is equivalent to applying their primal formulations using $\widetilde{\bPhi}$. Thus, a dual solution with $n$ pairwise comparisons can be similarly obtained via the corresponding primal domain optimization. Generally, the $n$ data points are sampled \emph{uniformly without replacement.} The computational complexity of this algorithm is $O(Nnk+n^3)$.

 \textbf{Ensemble Nystr\"{o}m} \cite{kumar2009ensemble} performs multiple smaller dimensional KPCA mappings, instead of a single large one. This can be done by dividing the $n$ samples into $m$ non-overlapping subsets to compute $m$ separate KPCAs. The resulting feature mappings are then scaled and concatenated. The computational complexity of this algorithm is $O(Nnk/m+n^3/m^2)\text{.}$ 

 
 \textbf{Nystr\"{o}m with Randomized SVD} \cite{li2010making,li2015large} speeds up the SVD in the Nystr\"{o}m algorithms via a randomized algorithm proposed in \cite{halko2011finding}. This reduces the computational complexity of Standard Nystr\"{o}m with rank reduction to $O(Nnk+n^2k+k^3)$.
 
 
 \section{Methodology}
 \vspace{-0.1pt}
 In this section, we explain the theory underlying our method and describe the components in our two stage procedure. Our method first approximates the negative margins for all training samples by utilizing standard kernel approximation methods, then trains a classifier with kernel based features extracted by using the chosen subset of samples, which we aptly name the \emph{support vectors}.
  
  \subsection{The Negative Margin Criterion}
  Kernel approximation methods primarily aim to create fast and accurate low rank approximations of the kernel matrix, even though approximating the kernel matrix itself is generally not an end goal. 
  While the Nystr\"{o}m variants need to sample a set that represents the training data well for a good approximation, the set of support vectors often constitute a small and unrepresentative subset of the data \cite{scholkopf2000support,friedman2001elements}. This observation motivates the selection method developed in this section.
  
  We shall first discuss why the \emph{negative margin} serves as a suitable criterion for the selection of the support vectors. As two examples, let us write the primal formulations of Soft-Margin SVM (SVM, left in \eqref{eq:primal}) and Ridge Regression, also known as Least-Squares SVM (RR, right in \eqref{eq:primal}),
  \begin{equation}
  \begin{split}
    \begin{aligned}
& \underset{\bw,\boldsymbol{\varepsilon},b}{\text{minimize}}
& & \frac{1}{2}\lnorm \bw \rnorm^2 + C\sum_{i=1}^{N} \varepsilon_i \\
& \text{subject to}
& & y_i(\bw^\top\bphi\lp\bx_i\rp+b) \geq 1-\varepsilon_i, \\
& & & \varepsilon_i \geq 0, \ \; \forall i.
	\end{aligned}
  \end{split}
  \qquad
  \begin{split}
    \begin{aligned}
& \underset{\bw,\boldsymbol{\varepsilon},b}{\text{minimize}}
& & \frac{1}{2} \lnorm \bw \rnorm^2 + \frac{1}{2\rho}\sum_{i=1}^{N} \varepsilon_i^2 \\
& \text{subject to}
& & y_i(\bw^\top\bphi\lp\bx_i\rp+b) = 1-\varepsilon_i, \ \; \forall i.
	\end{aligned}
  \end{split}
  \label{eq:primal}
\end{equation}
where $y_i \in \{-1,+1\}$. We can then introduce the dual coefficients $\{\alpha_i\}_{i=1}^{N}$ and derive the following optimality conditions for both models from the Karush-Kuhn-Tucker (KKT) conditions,
 \begin{equation}
 \begin{split}
 \bw = \sum_{i=1}^{N} y_i \alpha_i \bphi\lp\bx_i\rp
 \end{split}
 \quad \text{and} \quad
 \begin{split}
  \sum_{i=1}^{N} y_i\alpha_i = 0.
 \end{split}
 \end{equation}
 Moreover, we obtain the following optimality conditions on SVM (left in \eqref{eq:dual}) and RR (right in \eqref{eq:dual}),
 \begin{equation}
 \begin{split}
 \begin{aligned}
 & \alpha_i = 0, & &\text{if}\; y_i(\bw^\top\bphi\lp\bx_i\rp+b)>1, \\
 & 0 \leq \alpha_i \leq C, & &\text{if}\; y_i(\bw^\top\bphi\lp\bx_i\rp+b)=1, \\
 & \alpha_i = C, & &\text{if}\; y_i(\bw^\top\bphi\lp\bx_i\rp+b)<1.
 \end{aligned}
 \end{split}
 \qquad
 \begin{split}
  \alpha_i = \frac{1}{\rho}\left(1-y_i(\bw^\top\bphi\lp\bx_i\rp+b)\right).
 \end{split}
 \label{eq:dual}
 \end{equation}

From these conditions, it is clear that larger values of the negative margin, $-y_i(\bw^\top\bphi\lp\bx_i\rp+b)$ lead to larger weights for the samples. For SVM, samples get $0$ weights if this value is less than $-1$ and constant weights if it is greater than $-1$. For RR, the weights increase linearly with negative margin. 

The optimality conditions of SVM further imply that an optimal margin classifier can be found in the subspace spanned by the high negative margin samples. Additionally, applying the Standard Nystr\"om algorithm to data can simply be viewed as restricting the solution of the resulting model to the subspace spanned by $n$ chosen samples in the feature space \cite{yang2012nystrom,rudi2015less}. Thus, by performing a Nystr\"om approximation with high negative margin samples, it is possible to restrict the solution without losing optimality. 

Due to the symmetry of the least-squares loss, high positive margin samples may also get large RR weights in absolute value. But, since margin maximizing models reduce the pairwise comparisons by ignoring such samples, we use the negative margin as a support vector selection criterion for the RR classifier as well. This leaves us with the problem of finding such samples efficiently.

 
\subsection{Approximating the Margins by Approximating the Kernel}  
To find the margin values, one would normally train a classifier with the full kernel matrix, which defeats the purpose of using low dimensional feature mappings. However, by approximating the kernel matrix, one can also approximate the margin values. In addition, the bounds on the approximations are quite tight for the Ridge Regression model, which also enjoys optimal learning bounds with Nystr\"{o}m features \cite{rudi2015less}. In the following, we give the bound from Cortes \textit{et al.} \cite{cortes2010impact},
  
\begin{prop} \label{prop:margin_bound}
Let $m(\bx,y)$ and $\widetilde{m}(\bx,y)$ be the margins for the sample $(\bx,y)$ produced by KRR before and after the kernel approximation, respectively. Define $\kappa>0$ such that $K(\bx,\bx)\leq \kappa$ and $\widetilde{K}(\bx,\bx) \leq \kappa$ for all $\bx \in \mathcal{X}$. Then the following inequality holds for all $(\bx,y) \in (\mathcal{X},\mathcal{Y})$,
\begin{equation}
\left|m(\bx,y)-\widetilde{m}(\bx,y)\right| \leq \frac{N\kappa}{\rho^2 }\lnorm \bK-\widetilde{\bK} \rnorm_2.
\label{eq:bound}
\end{equation}
\end{prop}

 This bound allows us to to approximate the margin values at low computational costs by exploiting standard kernel approximation techniques. Quality of the margin approximation depends on the quality of the kernel approximation. Hence, in order to estimate the margin for all the samples, we can first approximate the kernel matrix using any of the methods described earlier. Afterwards, support vectors can be selected based on the approximate values of the negative margin. 
 
 \subsection{Nystr\"{o}m Kernel Ridge Regression Model}
 
 \begin{algorithm}[t]
\caption{Nystr\"{o}m Kernel Ridge Regression with support vector selection}
\textbf{Input:} Training data: $(\bX,\by)$; model parameters: $\rho$, $n_0$, $k_0$, $n_f$, $k_f$; and kernel parameters.
\begin{algorithmic} 
	\State \textbf{1.} Compute a $k_0$-dimensional approximate kernel feature map $\widetilde{\bPhi}$ using a Nystr\"{o}m variant with $n_0$ uniformly sampled data points.
    \State \textbf{2.} Train RR using the approximate feature map $\widetilde{\bPhi}$, by solving (\ref{eq:RR}).
    \State \textbf{3.} For $i \in \{1,\ldots,N\}$, compute the negative margin $-y_i\left(\bw_{y_i}^\top\bphi\lp\bx_i\rp+b_{y_i}\right)$, where $(\bw_{y_i},b_{y_i})$ are the parameters of the binary classifier that separates the class $y_i$ from the rest.
    \State \textbf{4.} Select $n_f$ data samples that maximize the negative margin.
    \State \textbf{5.} Compute the $k_f$-dimensional kernel feature map $\widetilde{\bPhi}_{sv}$ using a Nystr\"{o}m variant with the $n_f$ selected data points.
    \State \textbf{6.} Train another RR using $\widetilde{\bPhi}_{sv}$. Turn the resulting classifier into the standard form via (\ref{eq:standard}).
\end{algorithmic}
\textbf{Output:} \textit{One vs. Rest} Kernel Ridge Regression classifier with $n_f$ support vectors.
\label{alg:simple}
\end{algorithm}

 For the selection of support vectors and training of the final classifier, we use the Ridge Regression model, due to its provable generalization properties \cite{rudi2015less,rudi2017generalization} and tight bounds on the approximate margin, as presented by Proposition \ref{prop:margin_bound} \cite{cortes2010impact}. From our experience, we also found it to be robust to the choice of hyper-parameter. Thus, it makes a good choice for our application. To generalize our support vector selection method to multi-class settings, we train a \textit{One vs. Rest} RR, then use the classifier that separates a sample's own class from the others to compute the negative margin.
 
 From the approximate feature map $\widetilde{\bPhi}$, the RR classifier is obtained by solving
 \begin{equation}
 \underset{\bW,\bb}{\text{minimize}} \left\| \widetilde{\bPhi}^\top\bW+\rone\bb^\top-\bY \right\|_F^2+\rho\left\|\bW\right\|_F^2
 \label{eq:RR}
 \end{equation}
 where $\bY \in \mathbb{R}^{N \times L}$ is the class indicator matrix, with $L$ being the number of classes.
 
 Since any Nystr\"{o}m feature map can be written as $\widetilde{\bPhi}=\widehat{\bA}^\top\bK(\indSet_n,:)$, $\exists \widehat{\bA} \in \mathbb{R}^{n \times k}$, after finding the optimal $\bW$ and $\bb$, the resulting hypothesis can be applied directly to a test kernel matrix via
 \begin{equation}
  h\lp\bX_{test}\rp = \bK_{test}^\top\widehat{\bA}\bW+\rone\bb^\top = \bK_{test}^\top\bA + \rone\bb^\top \text{.}
 \label{eq:standard}
 \end{equation}
 This is simply a multi-class variant of the Support Vector Machine and is much cheaper to compute when $L \ll k$, as $\bA \in \mathbb{R}^{n \times L}$. Notice that immaterial which Nystr\"om variant or value of $k$ is used, $\bK_{test}$ has size $n \times N_{test}$, that is, the final classifier has $O(n)$ complexity, which is solely dependent on the number of samples used to compute the feature mapping (i.e., the support vectors).
 
 

\subsection{Summary}
Our method consists of two stages. We train an approximate kernel based classifier in the first stage. In the second stage, we select support vectors based on the approximate negative margins and train a classifier using only pairwise comparisons with selected samples. The Nystr\"{o}m methods described earlier form the backbone of these stages. 

The overall methodology is provided in Algorithm \ref{alg:simple}. Steps 1--2 form the first stage by training RR on features obtained from standard kernel approximation techniques. Steps 3--6 form the second stage by obtaining approximate margin values from the classifier in the first stage, selecting the support vectors, then training RR on features obtained from supervised kernel approximation. The first stage of the algorithm corresponds to training a standard, unsupervised Nystr\"{o}m RR model. Consequently, the novelty of our method comes from the second stage.

When $n_0 \approx n_f$, $k_0 \approx k_f$, steps 1--2 and 5--6 of the algorithm have similar computational costs, which depend on the Nystr\"{o}m variant used. Steps 3--4 add a combined computational cost of $O(Nk_0+N\log N)$, which is negligible when $\log N \ll n_0 k_0$. Thus, Algorithm \ref{alg:simple} has the same overall computational complexity as RR with standard Nystr\"{o}m techniques.


\section{Experiments}
\subsection{Experimental Setup}
\begin{table}[t]
  \caption{Summary of the datasets used in the experiments, $\gamma$ is the RBF kernel parameter.}
  \vspace{2mm}
  \centering
  \begin{tabularx}{\linewidth}{l Y Y Y Y Y}
  \toprule
  Dataset & \# Features & \# Training & \# Testing & \# Classes & $\gamma$ \\
  \midrule
  USPS \cite{hull1994database} & $256$ & $7291$ & $2007$ & $10$ & $0.01$ \\
   HAR \cite{anguita2013public} & $561$ & $7352$ & $2947$ & $6$ & $0.01$ \\
   Letter \cite{frey1991letter} & $16$ & $15000$ & $5000$ & $26$ & $1.0$ \\ 
   COD-RNA \cite{uzilov2006detection} & $8$ & $49451$ & $97824$ & $2$ & $1.0$ \\
   MNIST \cite{lecun1998gradient} & $784$ & $60000$ & $10000$ & $10$ & $0.01$ \\
   Buzz \cite{kawala2015prediction} & $77$ & $120000$ & $20707$ & $2$ & $0.001$ \\
   \bottomrule
  \end{tabularx}
  \label{tab:Datasets}
\end{table}

The datasets used in our experiments are summarized in Table \ref{tab:Datasets}. All the datasets, except for Buzz data, had default train/test splits at LibSVM \cite{chang2011libsvm} or UCI \cite{bache2013uci} repositories. The Buzz data was randomly split into train/test sets for each independent experiment. Training and testing sets of COD-RNA had duplicate entries, which are removed \emph{a priori}. We scaled the original features to be in $[0,1]$ and used RBF kernels, i.e.,
$K(\bx_i,\bx_j) = \exp( -\gamma \left\| \bx_i-\bx_j \right\|_2^2 )$, for all the experiments.
 
We performed 30 randomized trials for each experiment. The ridge $\rho$ was set to $10^{-5}$, as we found it to be a good value across all the datasets. For the Standard and Ensemble Nystr\"{o}m methods, we use $k=n$ to exploit the full space spanned by the chosen subset of samples.
We use $m=5$ for the Ensemble Nystr\"{o}m. For Nystr\"{o}m with Randomized SVD, we use $k=n/2$ for the computational gain. We set the oversampling parameter of Randomized SVD to $10$ and the power parameter to $2$.

We ran two sets of evaluations. (1) We evaluate the ability of Support Vector Selection to improve Standard Nystr\"{o}m. (2) We demonstrate the improvement of our method with other Nystr\"{o}m variants.

\begin{figure*}[t]
\begin{minipage}[b]{0.329\linewidth}
  \centering
  \centerline{\includegraphics[width=4.95cm]{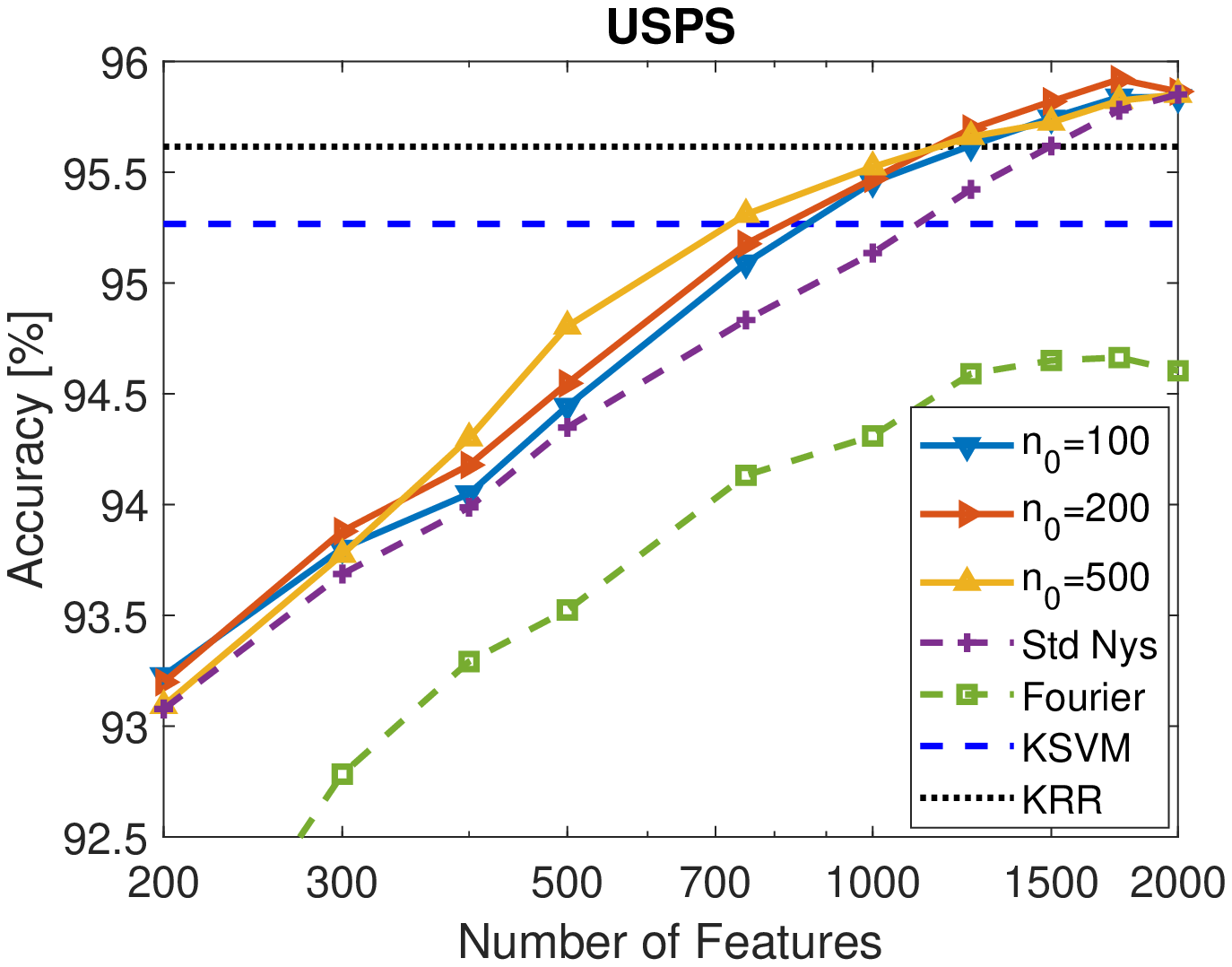}}
\end{minipage}
\hfill
\begin{minipage}[b]{0.329\linewidth}
  \centering
  \centerline{\includegraphics[width=4.95cm]{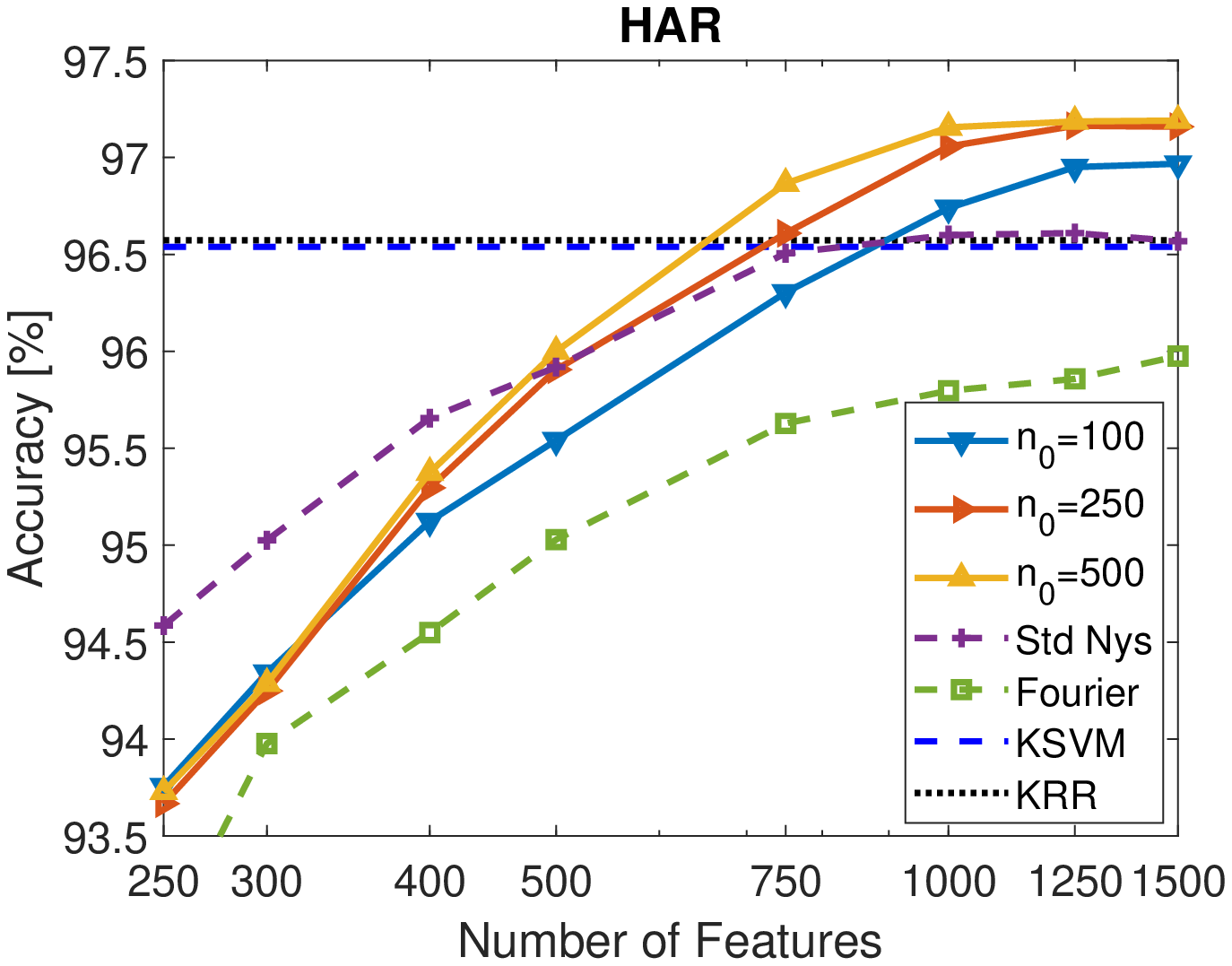}}
\end{minipage}
\hfill
\begin{minipage}[b]{0.329\linewidth}
  \centering
  \centerline{\includegraphics[width=4.95cm]{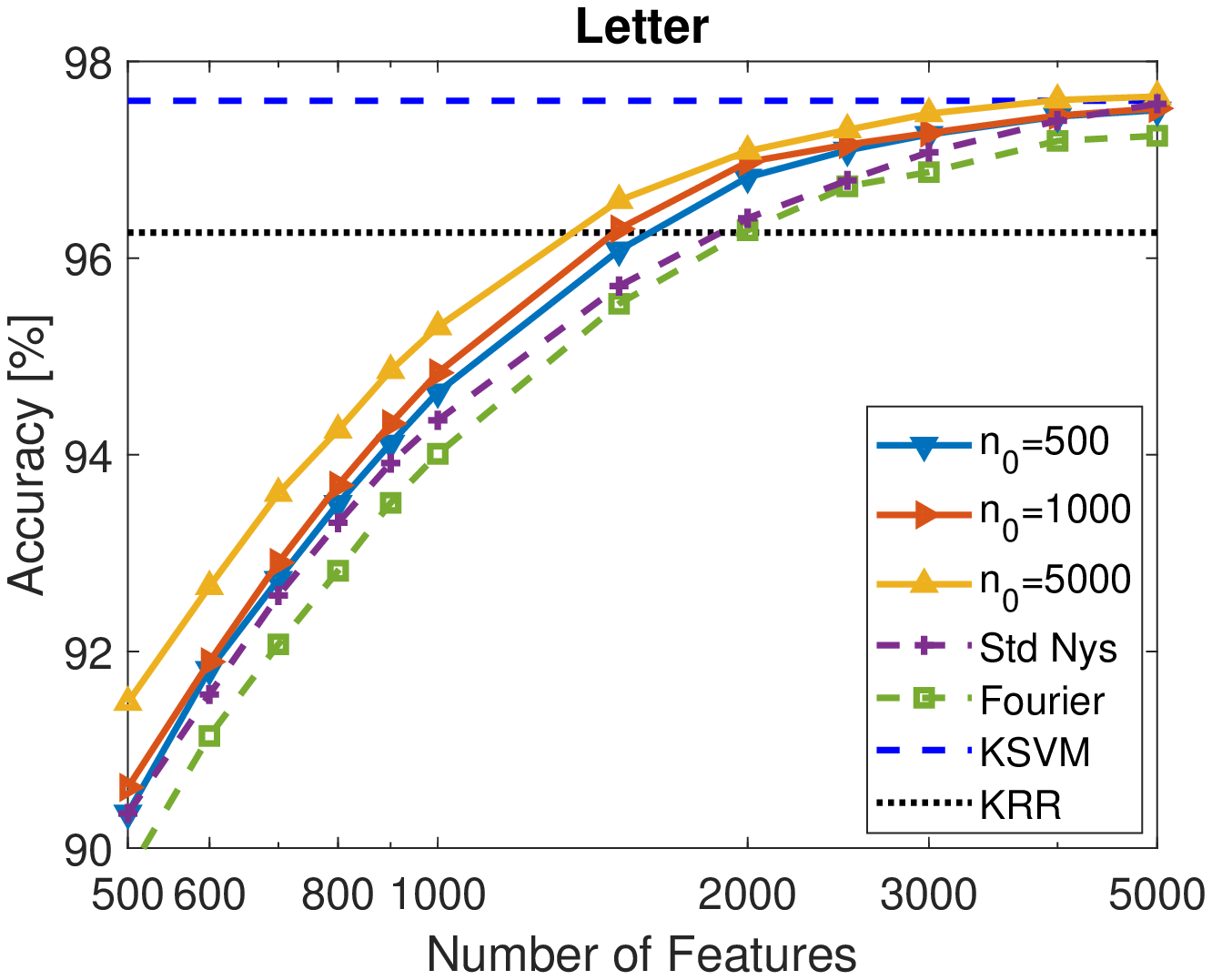}}
\end{minipage}
\begin{minipage}[b]{0.329\linewidth}
  \centering
  \centerline{\includegraphics[width=4.95cm]{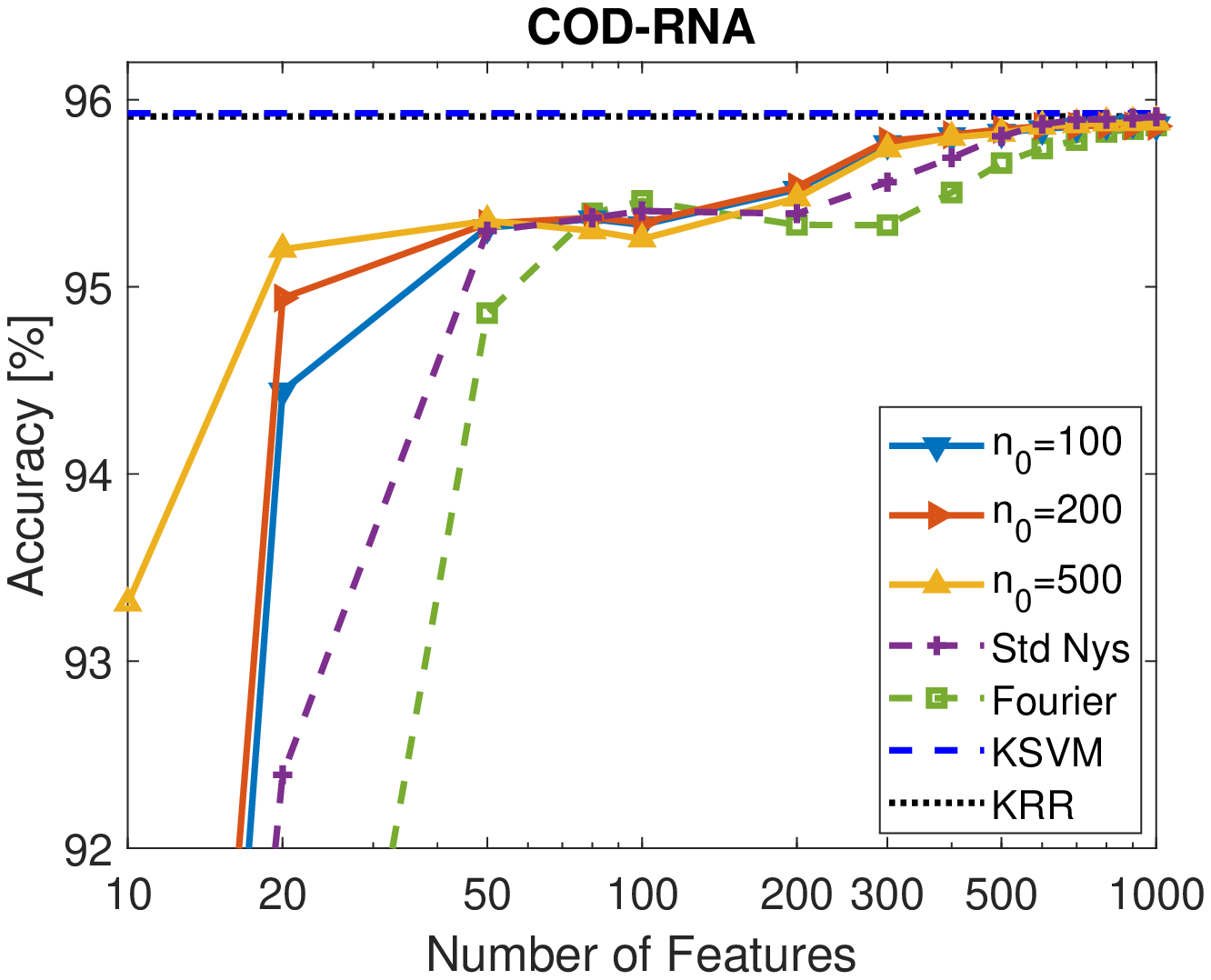}}
\end{minipage}
\hfill
\begin{minipage}[b]{0.329\linewidth}
  \centering
  \centerline{\includegraphics[width=4.95cm]{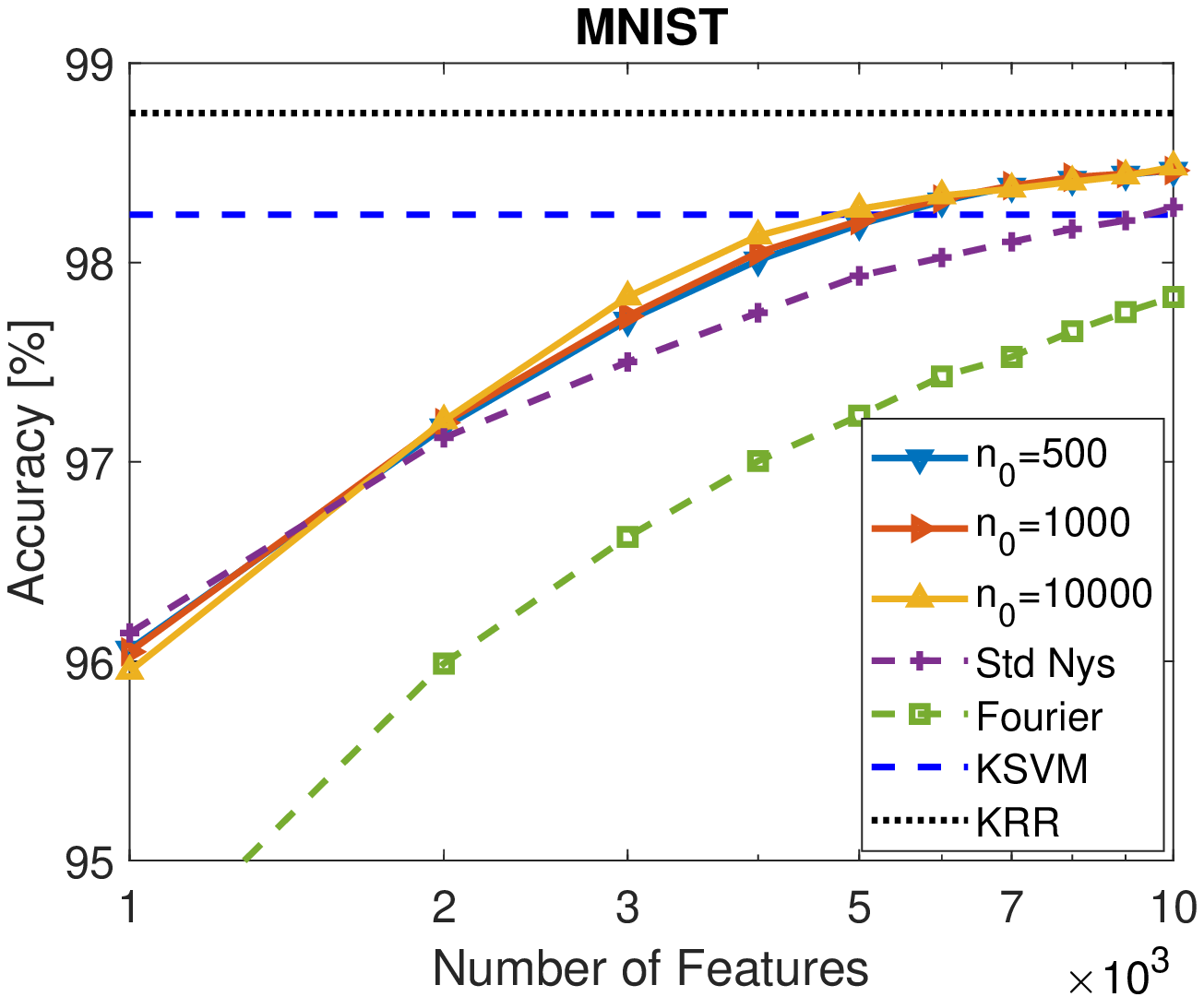}}
\end{minipage}
\hfill
\begin{minipage}[b]{0.329\linewidth}
  \centering
  \centerline{\includegraphics[width=4.95cm]{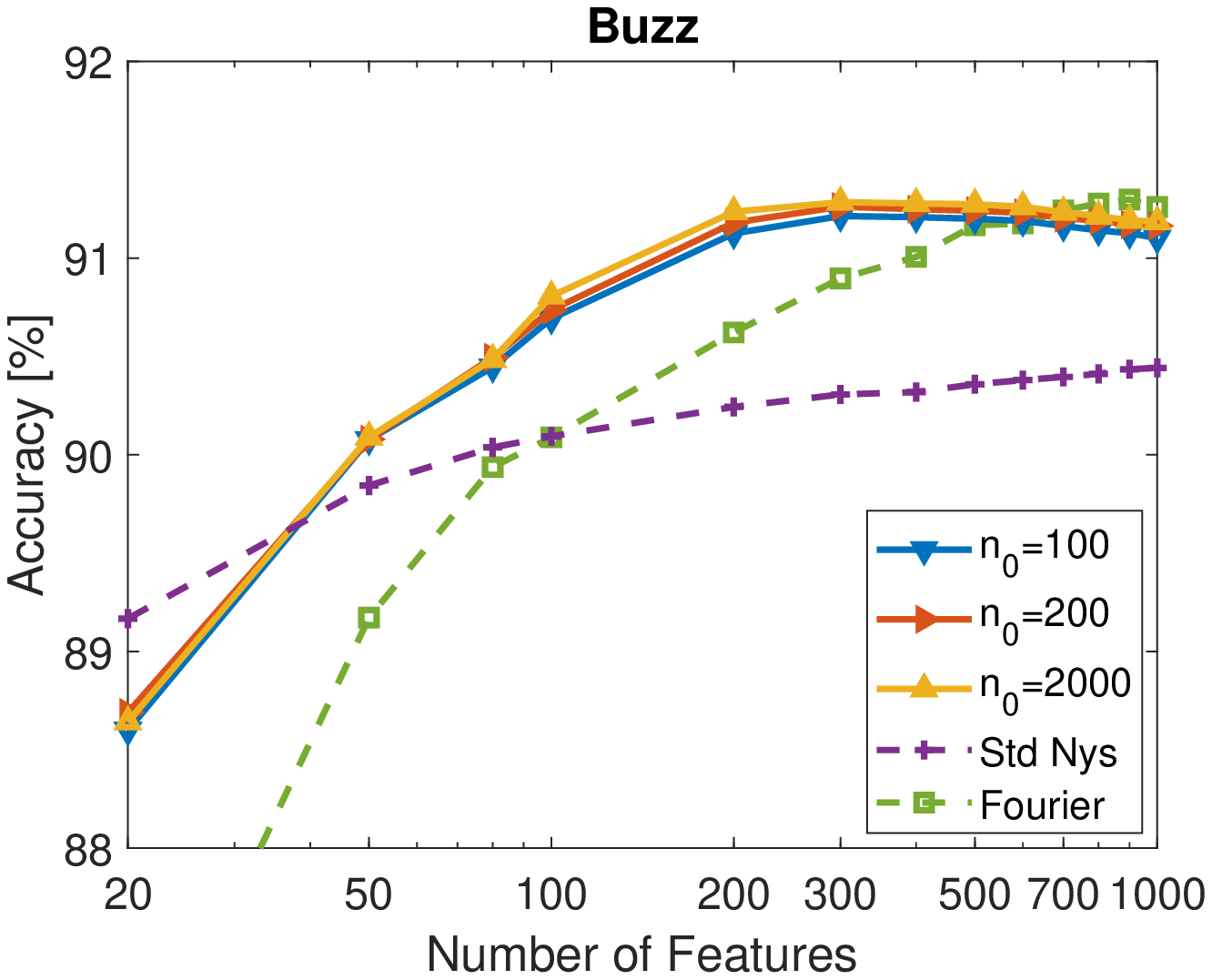}}
\end{minipage}
\caption{Number of features ($n_f$) vs. prediction accuracy. Accuracies of the final classifier generally increase with the initial approximation rank $n_0$. Supervised Nystr\"{o}m features outperform the Standard Nystr\"{o}m and Random Fourier features for various dimensions across the 6 datasets.
}
\label{fig:Results}
\end{figure*}

\subsection{Evaluation of Support Vector Selection with Standard Nystr\"{o}m}
We first demonstrate the success of support vector selection with varying degrees of the kernel approximations. We set $k_f=n_f$, $k_0=n_0$ in Algorithm \ref{alg:simple}, with $n_0$ controlling the quality of the margin approximations during support vector selection, and $n_f$ controlling the number of support vectors in the final model. The results are provided in Figure \ref{fig:Results}, where the Standard Nystr\"{o}m without support vector selection is denoted by \emph{Std Nys}. We include comparisons with full-dimensional
Kernel SVM (\emph{KSVM}) and Kernel RR (\emph{KRR}), except for the Buzz dataset, whose kernel matrix size is too big for the memory. We also include the performances of Random Fourier Features (\emph{Fourier}) \cite{rahimi2008random}.

First, the results show that the accuracy enhancement from support vector selection generally improves with the increase in the quality of the kernel matrix approximation, which agrees with Proposition \ref{prop:margin_bound}. Furthermore, we found that even with low approximation rank ($n_0$), support vector selection can yield notable improvement. Noticeably on MNIST, with $n_0=500$, we obtain $0.28\%$ accuracy increase over the Standard Nystr\"{o}m, but with $n_0=10000$, the marginal accuracy gain is at most $0.12\%$. Similarly for Buzz data, the difference in accuracies between $n_0=100$ and $n_0=2000$ is at most $0.15\%$ \textendash{} though $n_0=100$ already yields $0.9\%$ increase. \emph{Hence, the results show that significant predictive performance improvement can be achieved by utilizing a small number of samples for support vector selection, namely, by exploiting very cheap approximations of the margin values}.

Second, across all datasets, the Nystr\"{o}m with support vector selection, i.e., the \emph{supervised Nystr\"{o}m method}, outperforms the Standard Nystr\"{o}m for all but a few dimensions. 
Particularly on COD-RNA, where the highest accuracy achieved is $95.9\%$,
selecting only $10$ and $20$ support vectors with $n_0=500$ produces the accuracies of $93.4\%$ and $95.2\%$, respectively. These are remarkably higher than $84.1\%$ and $92.4\%$, the performances of the Standard Nystr\"{o}m with $10$ and $20$ dimensions.

Finally, the results demonstrate that \emph{with support vector selection, Nystr\"om methods can reach or exceed the predictive performances of full-dimensional mappings at even lower dimensions.} 
Particularly, on USPS, HAR and Letter, our method outperforms the full-dimensional \emph{KRR} at $1250$, $750$ and $1500$ dimensions, respectively, while the Standard Nystr\"om does not. Likewise, our method reaches the \emph{KSVM} performance faster on all the datasets, and outperforms \emph{KSVM} on USPS, HAR and MNIST by $0.65\%$, $0.65\%$ and $0.27\%$, using $1750$, $1250$ and $10000$ dimensions, respectively.

\begin{figure*}[t]
\begin{minipage}[b]{0.329\linewidth}
  \centering
  \centerline{\includegraphics[width=4.95cm]{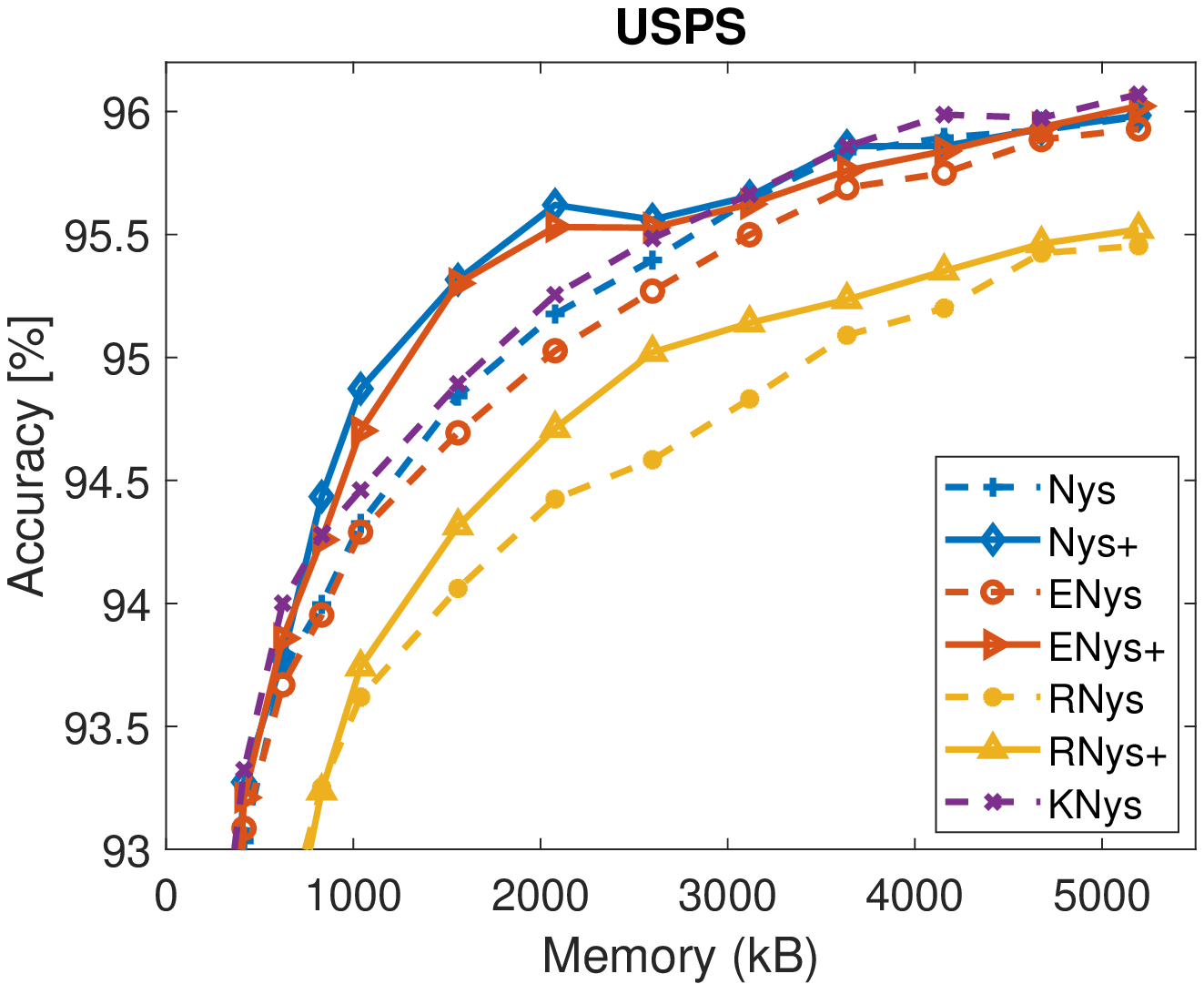}}
  \end{minipage}
\hfill
\begin{minipage}[b]{0.329\linewidth}
  \centering
  \centerline{\includegraphics[width=4.95cm]{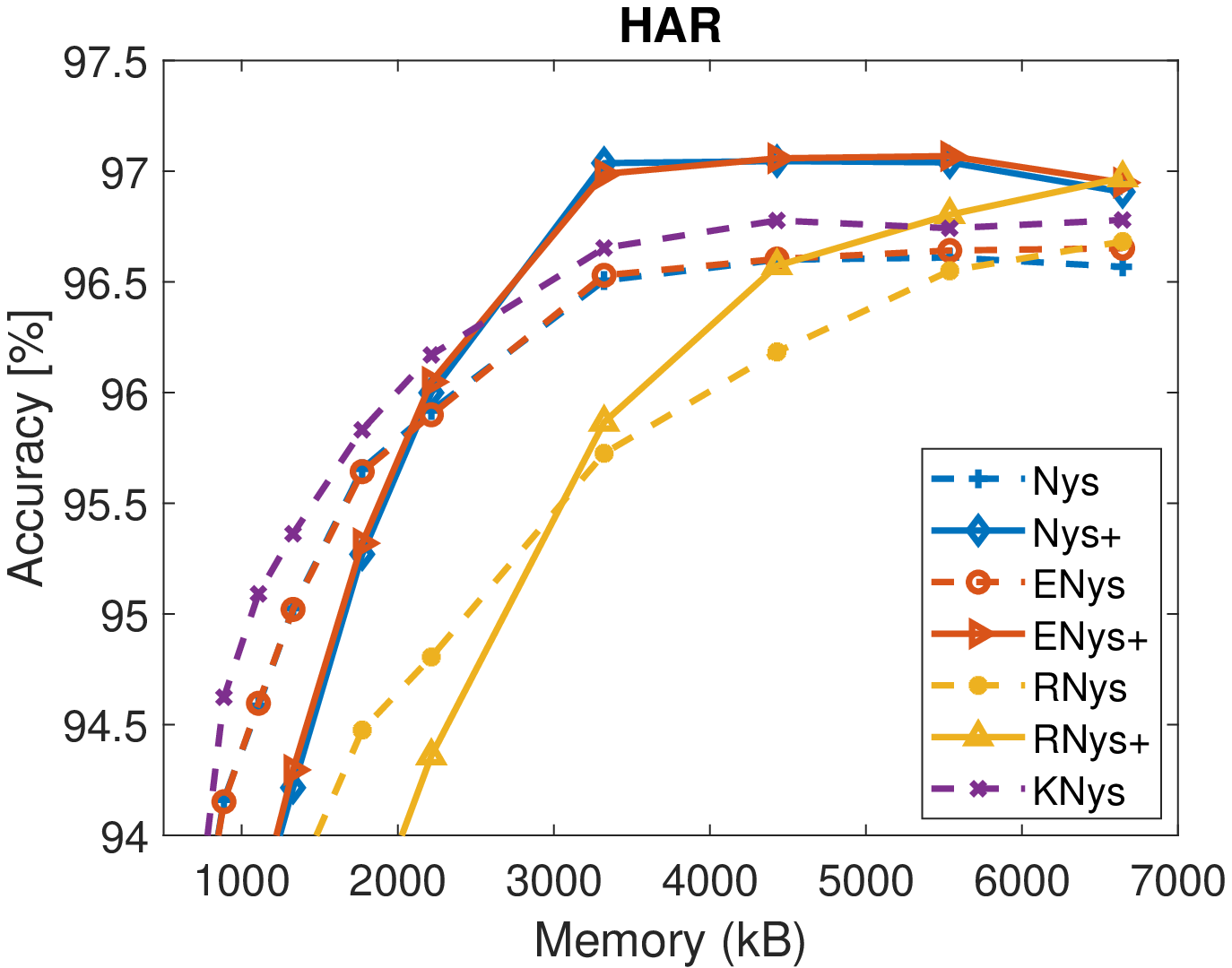}}
\end{minipage}
\hfill
\begin{minipage}[b]{0.329\linewidth}
  \centering
  \centerline{\includegraphics[width=4.95cm]{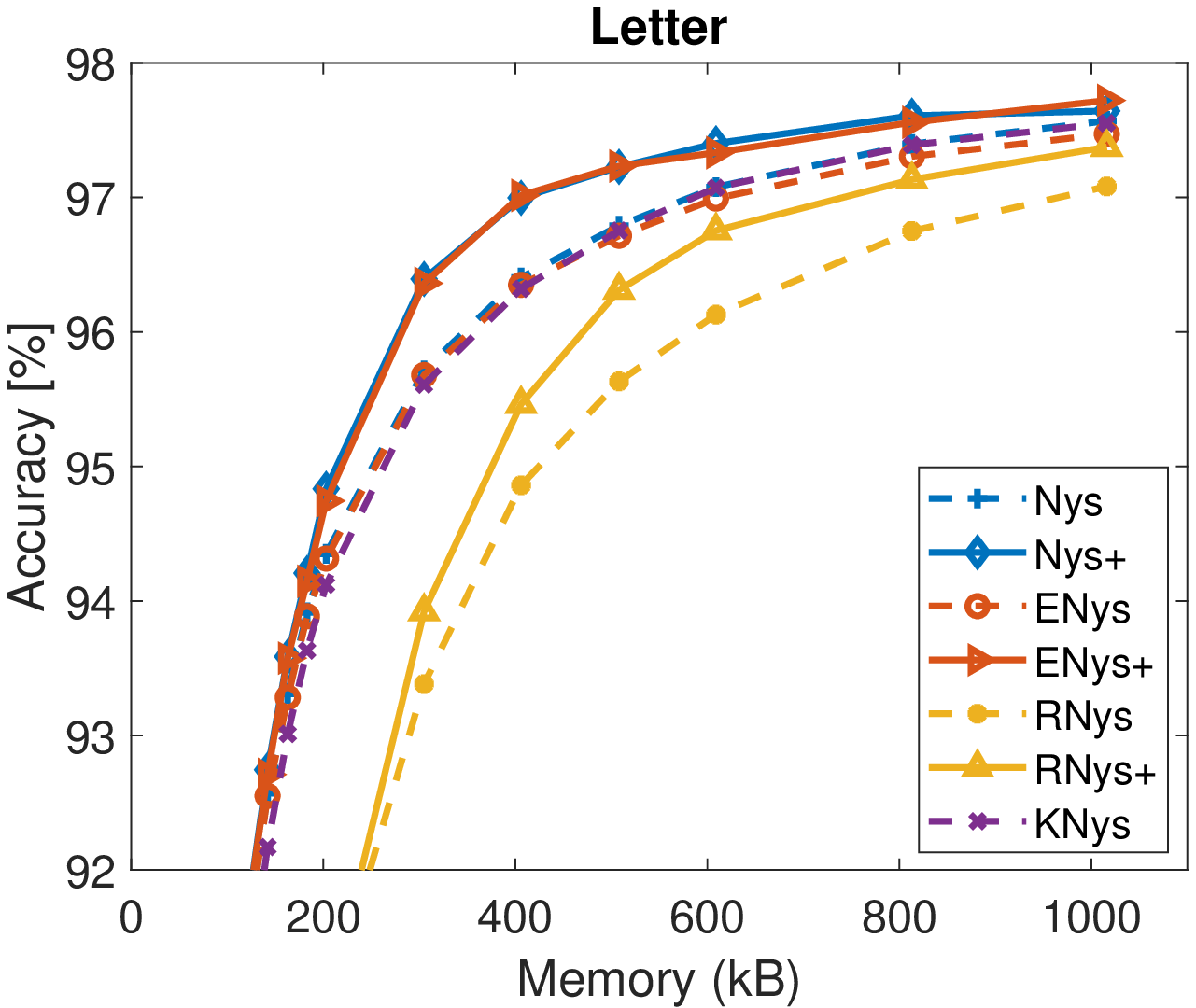}}
\end{minipage}
\begin{minipage}[b]{0.329\linewidth}
  \centering
  \centerline{\includegraphics[width=4.95cm]{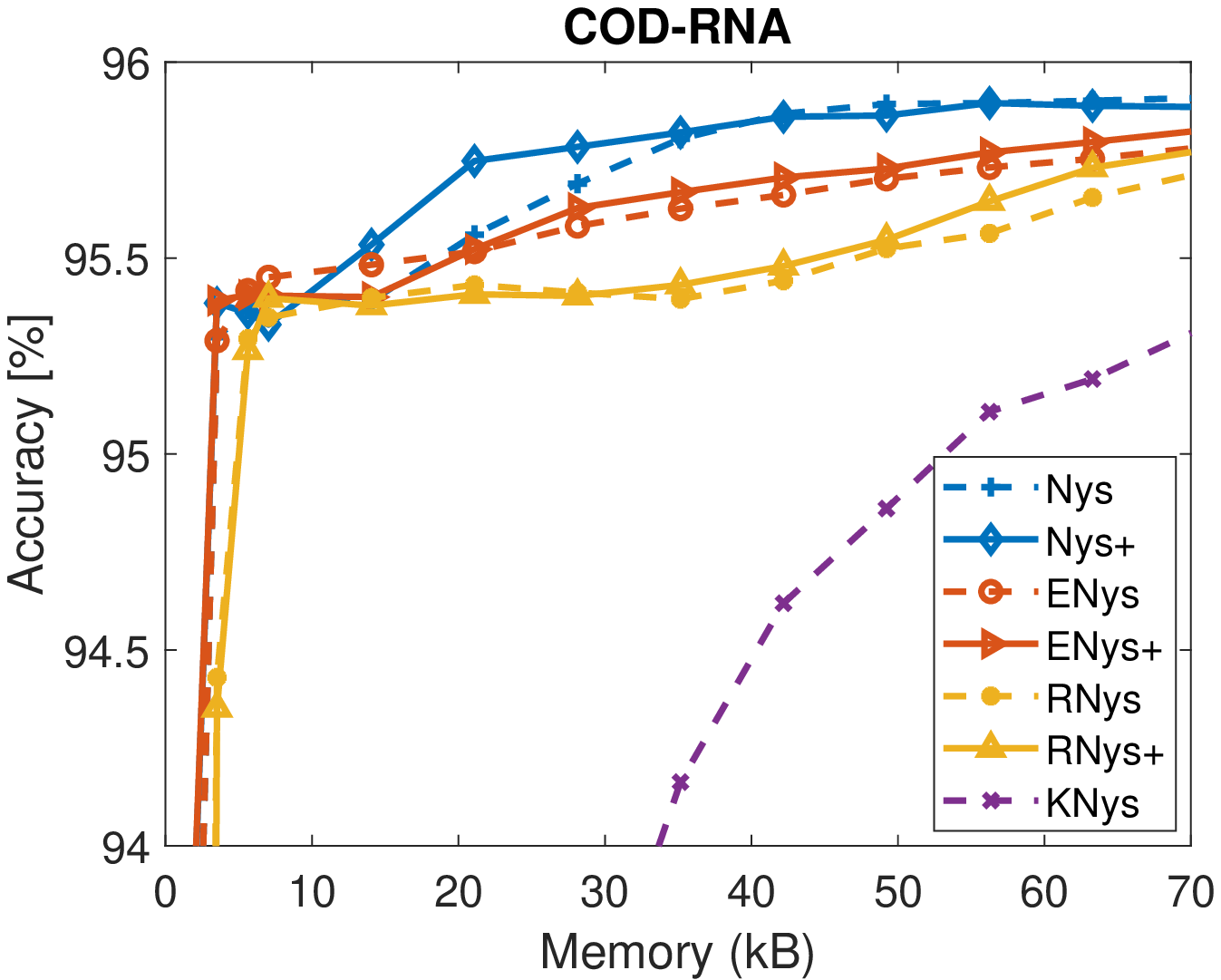}}
\end{minipage}
\hfill
\begin{minipage}[b]{0.329\linewidth}
  \centering
  \centerline{\includegraphics[width=4.95cm]{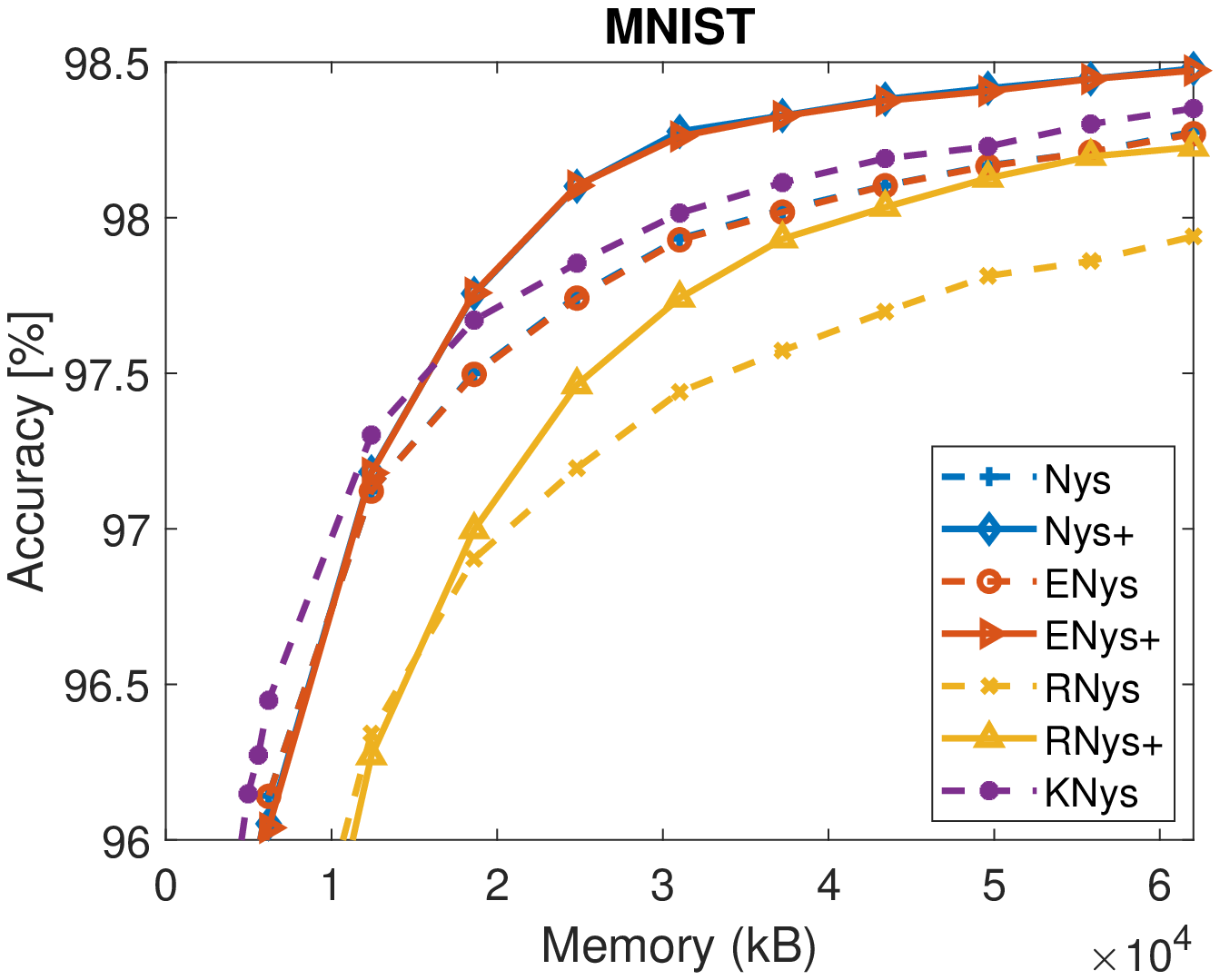}}
\end{minipage}
\hfill
\begin{minipage}[b]{0.329\linewidth}
  \centering
  \centerline{\includegraphics[width=4.95cm]{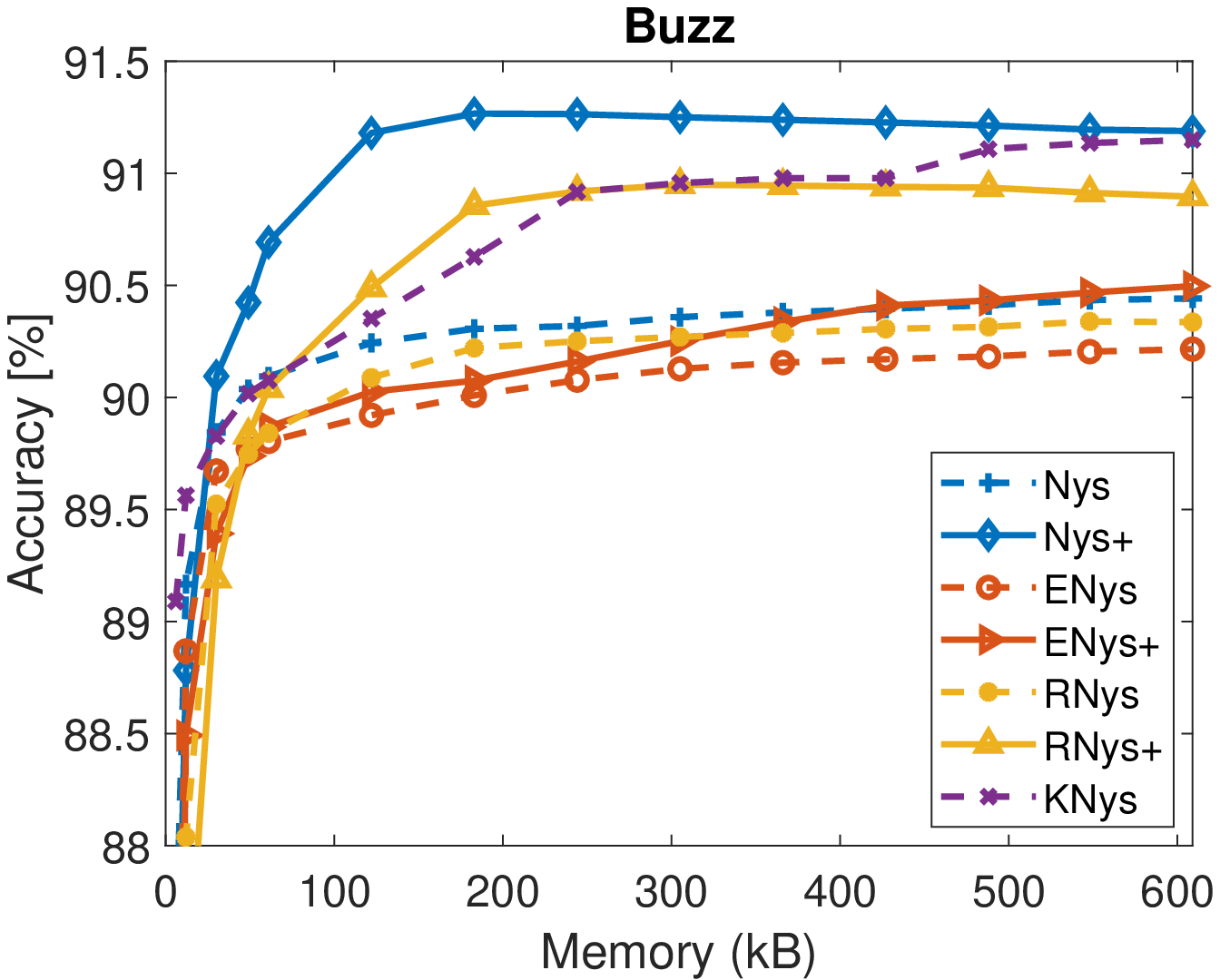}}
\end{minipage}
\caption{Size of the classifier model vs. prediction accuracy. For the best accuracy/model size trade-off, the supervised Standard Nystr\"{o}m (\emph{Nys+}) is the overall best method to use.}
\label{fig:Results2}
\end{figure*}

\begin{figure*}[t]
\begin{minipage}[b]{0.329\linewidth}
  \centering
  \centerline{\includegraphics[width=5.0cm]{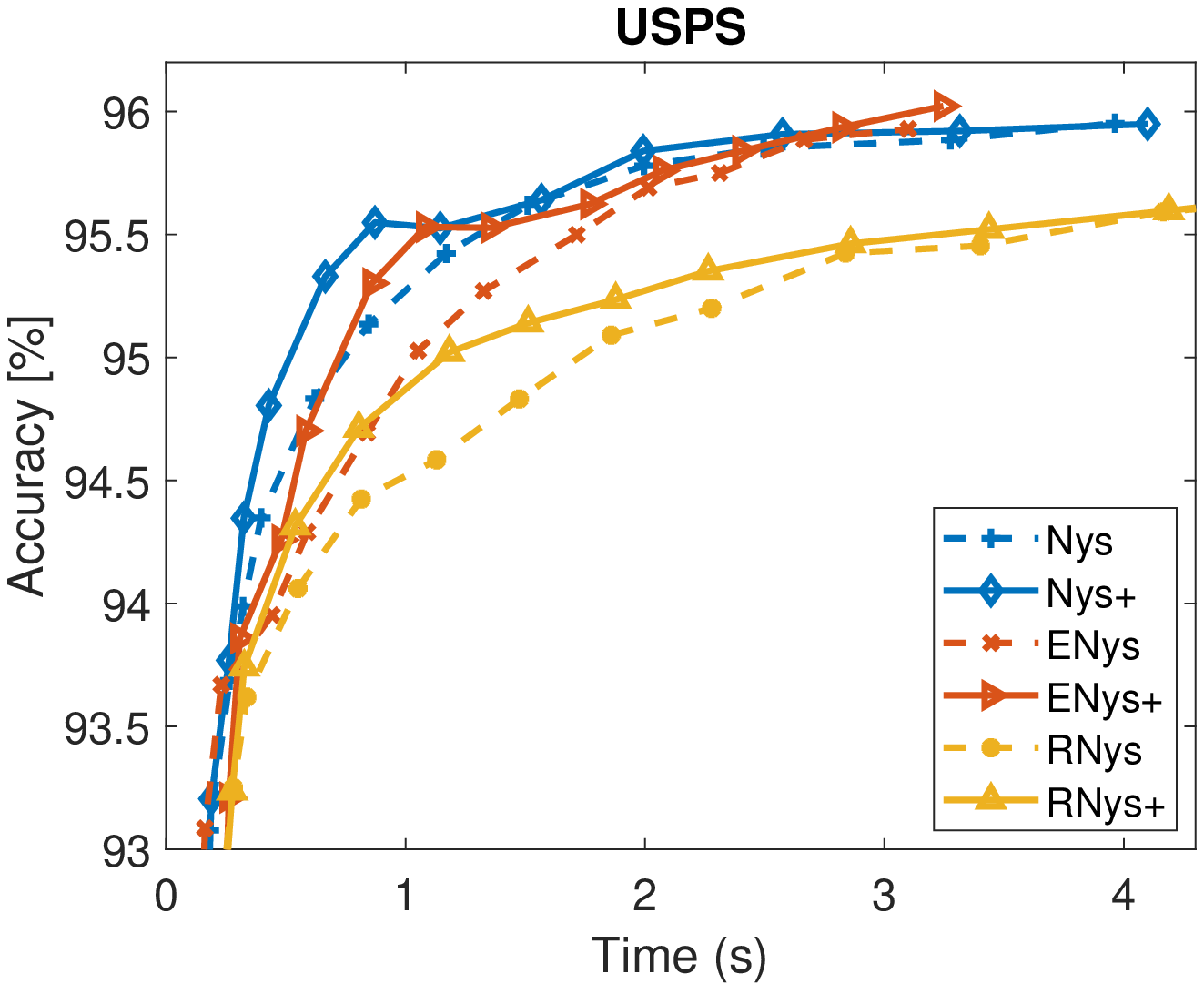}}
\end{minipage}
\hfill
\begin{minipage}[b]{0.329\linewidth}
  \centering
  \centerline{\includegraphics[width=4.95cm]{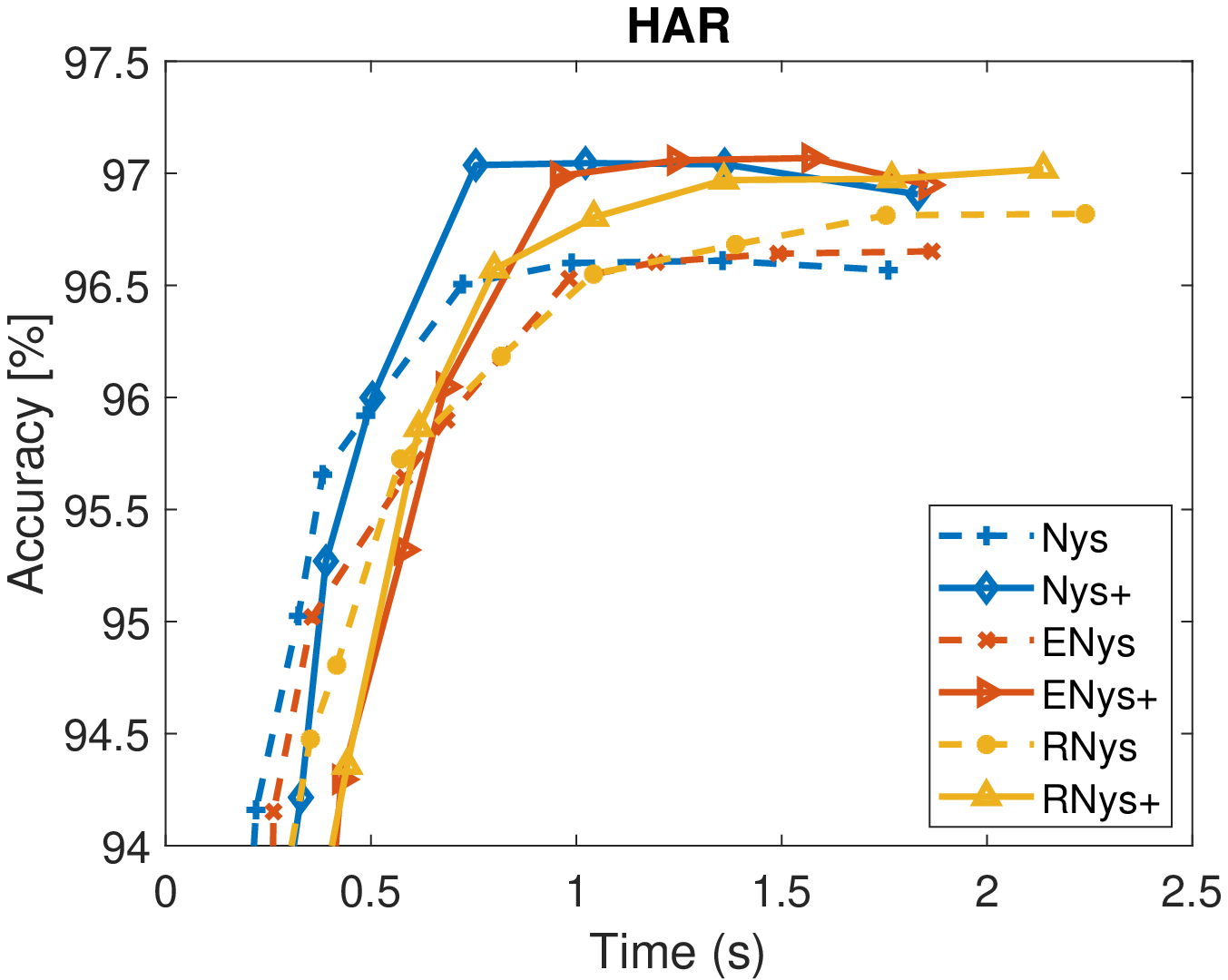}}
\end{minipage}
\hfill
\begin{minipage}[b]{0.329\linewidth}
  \centering
  \centerline{\includegraphics[width=4.95cm]{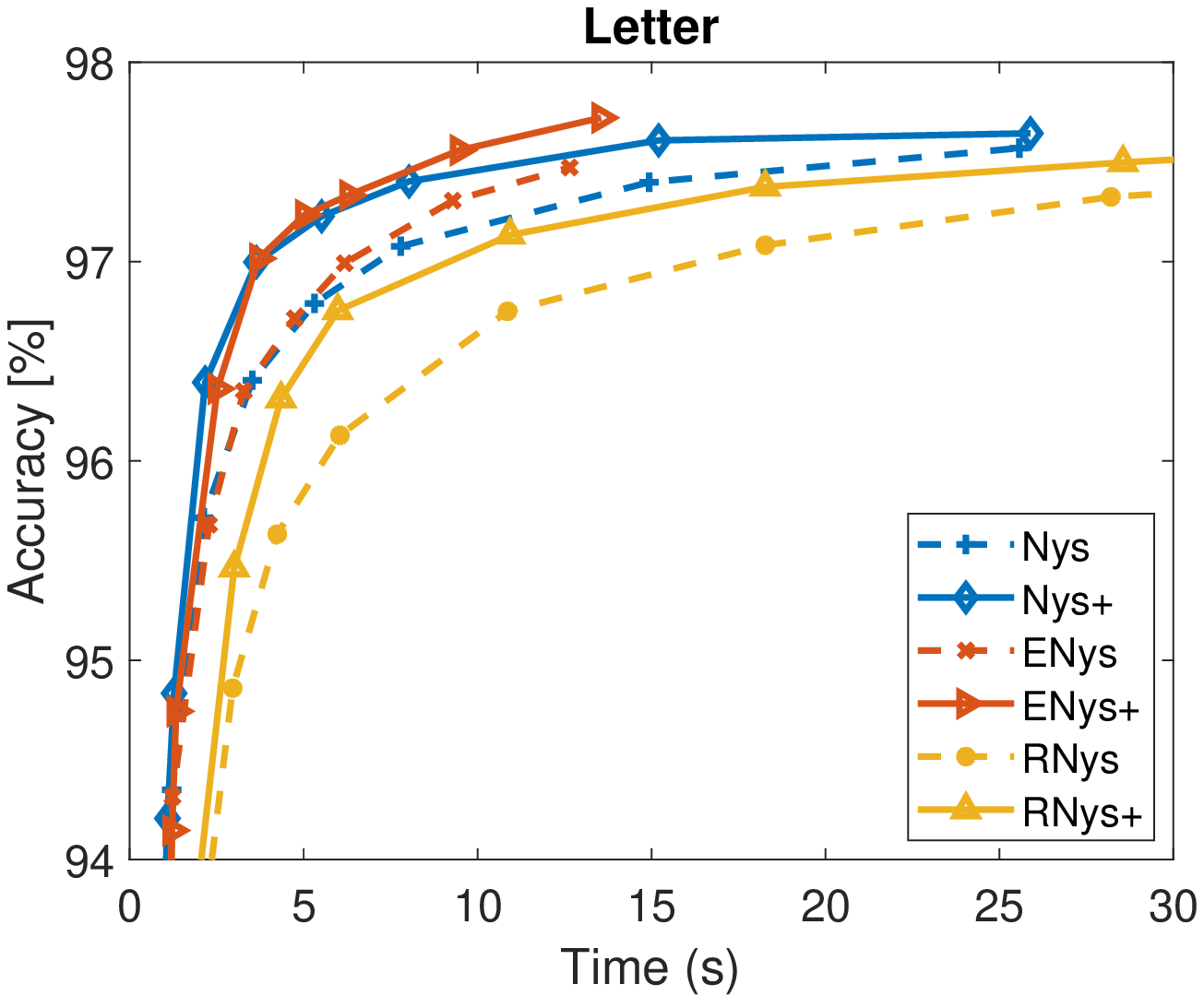}}
\end{minipage}
\begin{minipage}[b]{0.329\linewidth}
  \centering
  \centerline{\includegraphics[width=4.95cm]{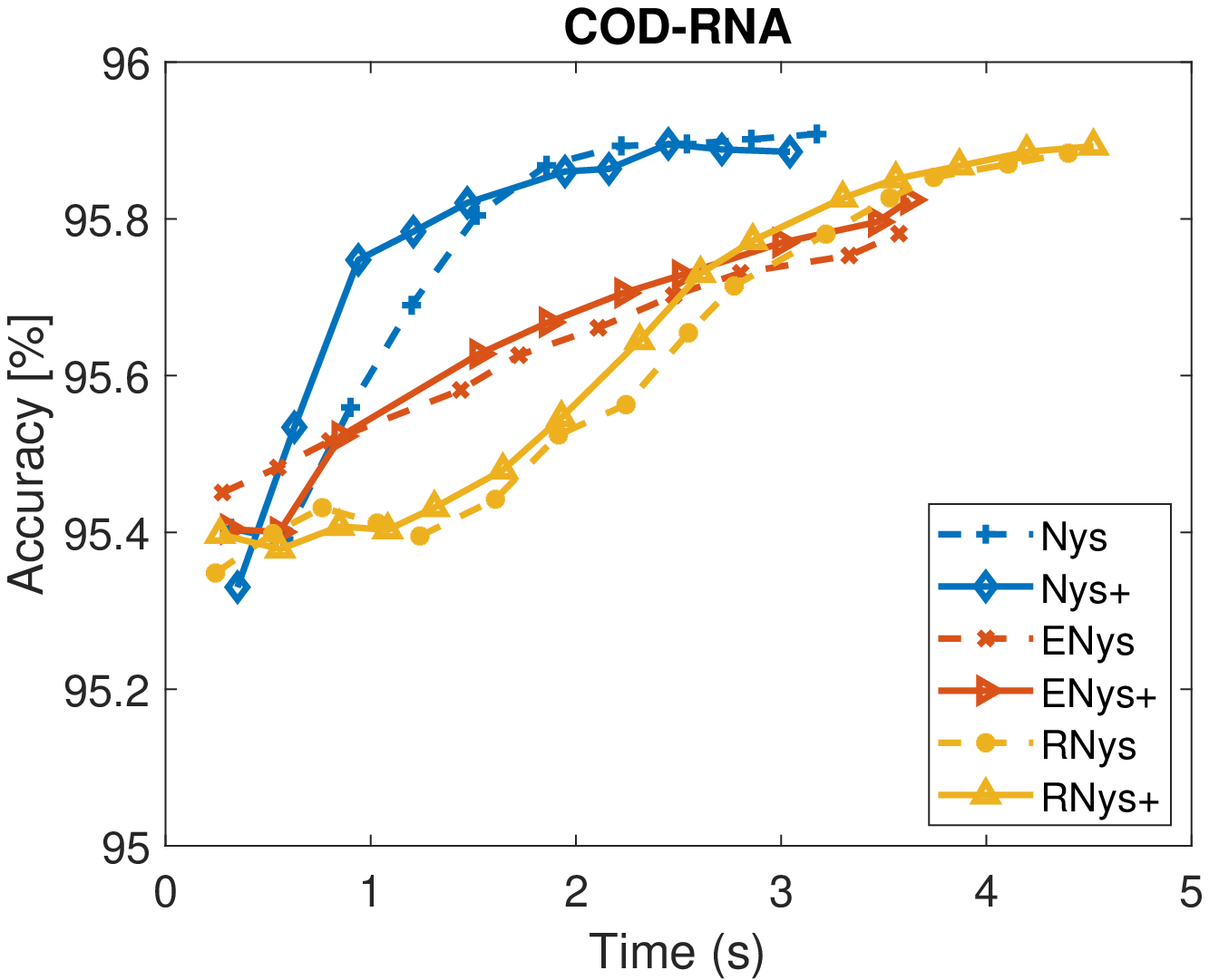}}
\end{minipage}
\hfill
\begin{minipage}[b]{0.329\linewidth}
  \centering
  \centerline{\includegraphics[width=4.95cm]{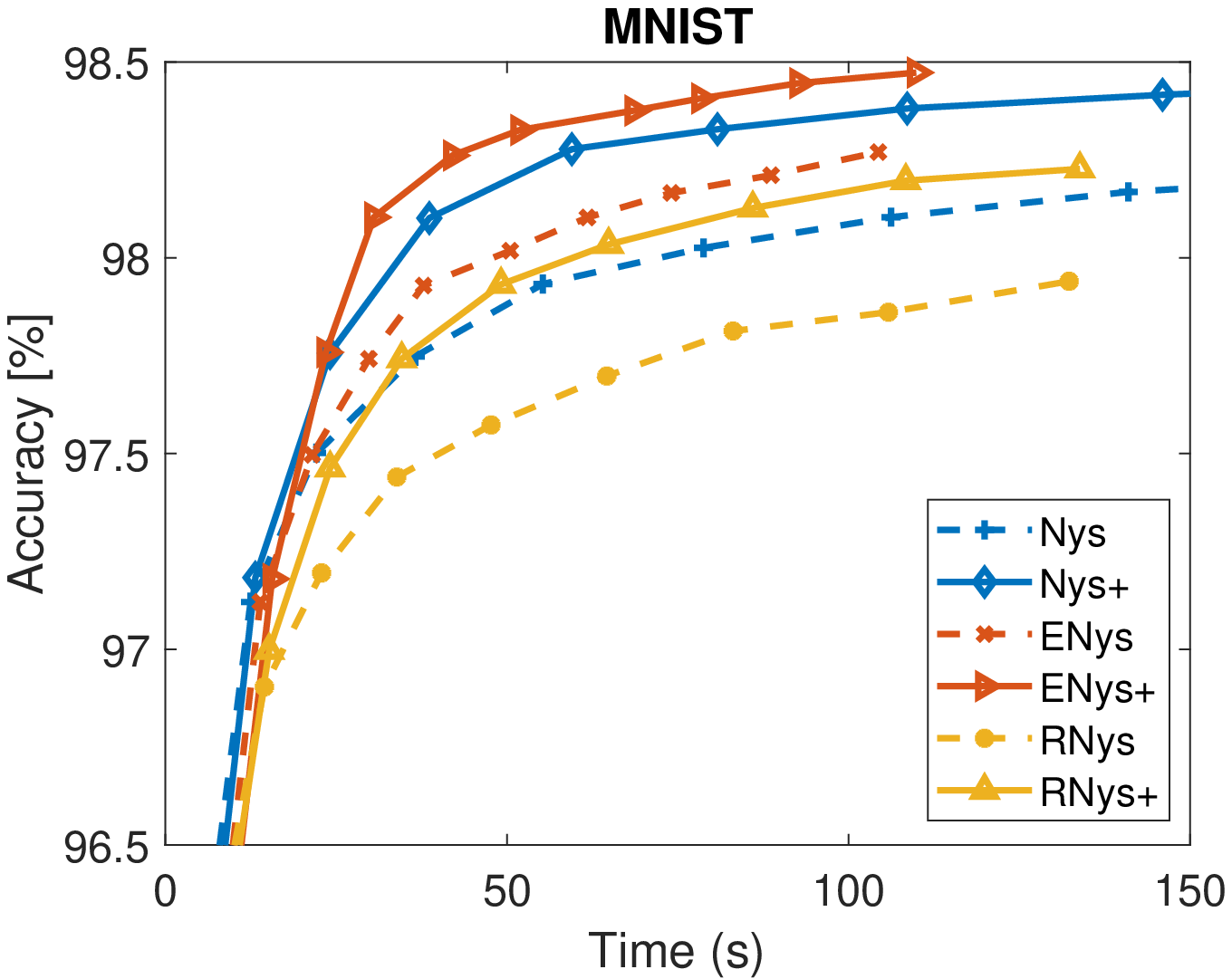}}
\end{minipage}
\hfill
\begin{minipage}[b]{0.329\linewidth}
  \centering
  \centerline{\includegraphics[width=4.95cm]{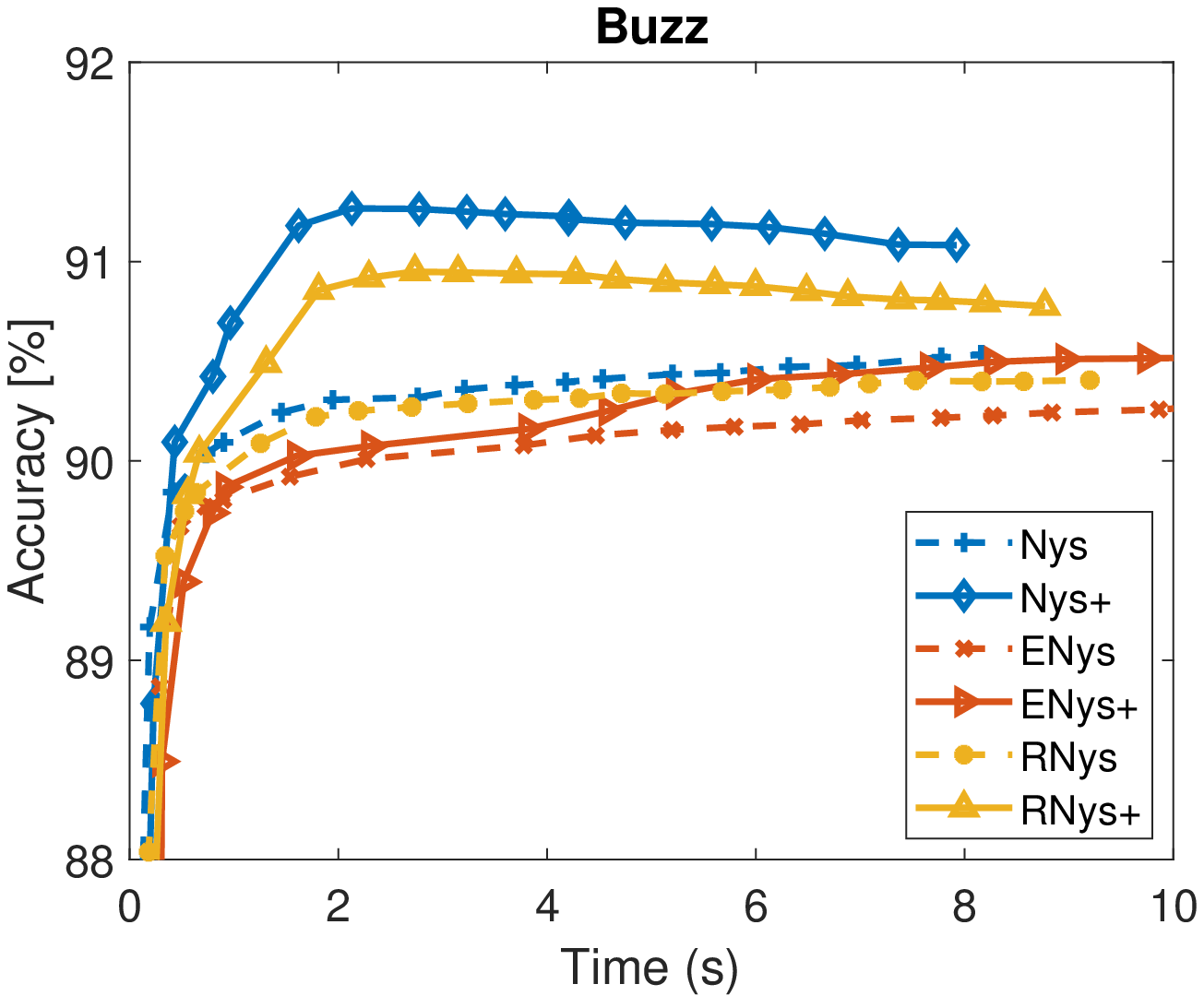}}
\end{minipage}
\caption{Training time vs. prediction accuracy. The supervised variant curves show \emph{the additional times} spent for support vector selection and training of the final classifier (steps 3--6 in Algorithm \ref{alg:simple}).}
\label{fig:Results3}
\end{figure*}

\subsection{Improvement Over Other Nystr\"{o}m Variants}
\vspace{-3pt}

In this section, we analyze the improvement of our support vector selection upon the Standard, Ensemble, and Randomized SVD variants of Nystr\"{o}m, denoted by \emph{Nys}, \emph{ENys} and \emph{RNys}, respectively. Although lower values of $n_0$ were shown to be viable in the previous section, we set $n_0=n_f$. This is a suitable scenario for training kernel machines under memory constraints. Since the two stages of our algorithm have similar computational costs under this setting, this roughly doubles the total training time over unsupervised Nystr\"om but does not change the overall complexity. As another comparison, we include the K-Means Nystr\"{o}m with the same settings as \cite{zhang2008improved}, denoted by \emph{KNys}. 



The model sizes and training times of RR classifiers with different Nystr\"{o}m variants are demonstrated in Figures \ref{fig:Results2} and \ref{fig:Results3}, respectively. \emph{Nys}, \emph{ENys}, \emph{RNys} and \emph{KNys} are the results without support vector selection (dashed curves), whereas \emph{Nys+}, \emph{Enys+} and \emph{RNys+} are the results with support vector selection (solid curves). Note that the solid curves (supervised Nystr\"{o}m variants) in Figure \ref{fig:Results3} show the times spent in the second stage (steps 3--6) of Algorithm \ref{alg:simple}. The first stage (steps 1--2) applies the same algorithms without support vector selection, the training times for which are shown by the dashed curves. We do not include K-Means Nystr\"{o}m in Figure \ref{fig:Results3}, as the computational overhead from k-means led to its consistent under-performance in terms of training time. The specific observations on each dataset are as follows.

\textbf{USPS:} All Nystr\"{o}m variants are significantly improved by support vector selection for a range of memory/time complexities, before accuracy begins to saturate.
Among all compared methods, \emph{Nys+} performs the best by outperforming
\emph{KNys} by $0.37\%$ and \emph{Nys} by $0.44\%$ at $2078$ kB model size.

\textbf{HAR:} All Nystr\"{o}m variants are significantly improved by support vector selection after a certain memory/time complexity is reached. \emph{Nys+} performs the best and outperforms \emph{KNys} by $0.39\%$ and \emph{Nys} by $0.51\%$ at $3322$ kB model size.

\textbf{Letter:} All Nystr\"{o}m variants are significantly improved by support vector selection for a wide range of memory/time complexities. \emph{ENys+} performs the best and outperforms \emph{KNys} and \emph{ENys} by more than $0.67\%$ at $406$ kB model size.

\textbf{COD-RNA}: \emph{Nys} is significantly improved for a range of complexities, before accuracy begins to saturate. \emph{ENys} and \emph{RNys} are consistently improved, but to a lesser extent. \emph{Nys+} performs the best and outperforms \emph{KNys} and \emph{Nys} by more than $0.19\%$ at $21$ kB model size, requiring $40\%$ less size to reach within one standard deviation of the saturation accuracy.

\textbf{MNIST}: All Nystr\"{o}m variants are significantly improved for a wide range of complexities. 
\emph{ENys+} performs the best and outperforms \emph{KNys} by $0.26\%$ and \emph{ENys} by $0.35\%$ at $406$ kB model size.

\textbf{Buzz}: All Nystr\"{o}m variants are significantly improved for a wide range of complexities. 
\emph{Nys+} performs the best and outperforms \emph{KNys} by $0.83\%$ and \emph{Nys} by $0.94\%$ at $122$ kB model size.

We observe from Figure \ref{fig:Results2} that among the unsupervised variants, \emph{KNys} is the best on USPS, HAR, MNIST and Buzz datasets for achieving high accuracy with the smallest model size. However, it is consistently outperformed by \emph{Nys+} after supervision is introduced.

Furthermore, Figure \ref{fig:Results3} shows that steps 1--2 and 3--6 of Algorithm \ref{alg:simple} cost roughly the same training times when $n_0=n_f$, $k_0=k_f$, and this setting suffices to give significant improvement. Hence, \emph{Algorithm \ref{alg:simple} can enhance Nystr\"{o}m variants without increasing their overall computational complexity.}

\subsubsection{Summary}
Our experimental observations support the claim that choosing the right solution subspace via support vector selection can be imperative for the success of scalable kernel based learning. More importantly, they illustrate that the negative margin criterion is suitable for selecting the data samples that span a good solution subspace.

In addition, support vector selection consistently improves the performances of Standard Nystr\"{o}m and Nystr\"{o}m with randomized SVD. These are also the variants that directly project the data into a subspace spanned by the chosen samples. Ensemble Nystr\"{o}m introduces scaling across subspace dimensions due to the lack of orthonormality between projection directions, so it does not conform fully to the main idea behind support vector selection. Nevertheless, Ensemble Nystr\"{o}m is also improved on all the datasets, and the improvements are very significant on data where scaling does not have an adverse effect on the performance.


\section{Discussion}

Although our method works well for classification on the 6 datasets, there are certain ways the support vector selection strategy can be improved and extended:
\begin{itemize}[leftmargin=0.5cm]
\item To ensure that successive support vectors add more information on datasets with highly similar/redundant samples, a dissimilarity, or orthogonality criterion can be employed as in \cite{ouimet2005greedy}, in addition to the negative margin. 
\item A mechanism can be added to remove the outlier support vectors before selection. A simple way to do this would be to apply a threshold on the negative margin.
\item The support vector selection method can be applied to regression tasks, by replacing the negative margin with the least-squares error as the selection criterion.
\end{itemize}
We leave these modifications to the algorithm, as well as application of our method with different Machine Learning models for future work. Nonetheless, we present possible extensions with K-Means Nystr\"{o}m in Appendix \ref{app:k-means}, and a scheme to deal with data redundancy in Appendix \ref{app:support_centroid}.


\section{Conclusion}
We have proposed a supervised sample selection methodology for Nystr\"{o}m methods to improve their predictive performances. Our selection method, inspired by the dual formulations of multiple machine learning models, successfully improves the classification performance obtained from Nystr\"{o}m variants, and leads to better classifiers in test time in terms of both accuracy and complexity. Moreover, our method allows this improvement to be achieved at a cost no more than that of training a classifier using standard Nystr\"{o}m techniques.

\subsubsection*{Acknowledgments}
This material is based upon work supported in part by the Brandeis Program of the Defense Advanced Research Project Agency (DARPA)
and Space and Naval Warfare System Center Pacific (SSC Pacific)
under Contract No. 66001-15-C-4068. We thank Prof. J. Morris
Chang and Prof. Pei-Yuan Wu for making this research possible.


\small
{\bibliographystyle{ieeetr}
\bibliography{refs}}

\normalsize
\begin{appendix}

\section{Nystr\"om Variants} \label{app:variants}
\subsection{Ensemble Nystr\"om}
 Ensemble Nystr\"{o}m \cite{kumar2009ensemble} performs multiple smaller dimensional KPCA mappings, instead of a single large one. For $m$ experts, each using non-overlapping subsets of $n'=n/m$ data samples, and with $k'=k/m$, this algorithm produces a rank-$k$ approximation of the kernel matrix given by
 \begin{equation}
  \widetilde{\bK} = \sum_{i=1}^{m} \mu_i \bC^{(i)} {\bB_{k'}^{(i)}}^{+} {\bC^{(i)}}^\top = \bC \blkdiag^+\left(\left\{\frac{1}{\mu_i}\bB_{k'}^{(i)}\right\}_{i=1}^m\right)\bC^\top\text{,}
 \end{equation}
 where $\{\mu_i\}_{i=1}^{m}$ are positive expert weights that add up to $1$\footnote{Computing the ensemble weights introduces additional overhead and lacks significant benefits for the task of classification, therefore the weights can be set to $\nicefrac{1}{m}$.}, $\blkdiag(\cdot)$ produces a block-diagonal matrix from its inputs, $\bC=[\bC^{(1)}\ \bC^{(2)}\ \cdots\ \bC^{(m)}]$, $\bC^{(i)} = \bK(:,\indSet_{n'}^{(i)})$, $\bB^{(i)}=\bK(\indSet_{n'}^{(i)},\indSet_{n'}^{(i)})$, and $\indSet_{n'}^{(i)}$ denotes the index set of $n'$ samples used by the $i^{th}$ expert. This approximation is equivalent to computing $m$ KPCAs over the non-overlapping subsets of samples, and concatenating the resulting feature mappings applied to the training data,
 \begin{equation}
 \widetilde{\bPhi} =  \concat\left(\left\{\sqrt{\mu_i}{\bSigma^{(i)}}^{-\nicefrac{1}{2}}{\bU^{(i)}}^\top{\bC^{(i)}}^\top \right\}_{i=1}^{m}\right)
 \end{equation}
 where $\concat(\cdot)$ is row-wise concatenation, and $\bB_{k'}^{(i)}=\bU^{(i)}{\bSigma^{(i)}}{\bU^{(i)}}^\top$ is the compact SVD of $\bB_{k'}^{(i)}$. 

Dividing the algorithm into $m$ smaller KPCAs reduces its computational complexity to $O(Nnk/m+n^3/m^2)\text{.}$ Unlike the Standard Nystr\"{o}m (i.e., $m=1$), however, the resulting feature projections will almost never be mutually orthonormal.

\subsection{Nystr\"{o}m with Randomized SVD}
 Nystr\"{o}m with Randomized SVD \cite{li2010making,li2015large} speeds up the SVD in the Nystr\"{o}m algorithms via a randomized algorithm proposed in \cite{halko2011finding}. This reduces the computational complexity of Standard Nystr\"{o}m with rank reduction to $O(Nnk+n^2k+k^3)$. It can also be applied to speed up the individual KPCAs in Ensemble Nystr\"{o}m, though we apply this method only to Standard Nystr\"om in this paper. In order to obtain a considerable speed up, $k$ needs to be much smaller than $n$. Although rank reduction further speeds up classifier training, it does not affect the complexity of the final model in test time, and generally does not enhance the predictive performance.

\subsection{K-Means Nystr\"{o}m}
 K-Means Nystr\"{o}m \cite{zhang2008improved} replaces uniform sampling with k-means. The representative set of samples are chosen to be the $n$ cluster centroids produced by k-means, which runs on a larger subset of the training data. It produces a rank-$k$ approximation of the kernel matrix given by $\widetilde{\bK}=\bC\bB_k^+\bC^\top$. In this case, $\bC$ is an $N \times n$ kernel matrix containing pairwise similarities between the training samples and the cluster centroids, while $\bB$ is an $n \times n$ kernel matrix containing pairwise similarities between just the cluster centroids. 
 
 This method is the same as computing KPCA using the cluster centroids and applying the resulting feature mapping to the training data. Running a fixed number of k-means iterations introduces additional overhead, but does not change the overall complexity over the Standard Nystr\"{o}m algorithm.

\section{K-Means Nystr\"{o}m Experiments} \label{app:k-means}
 \subsection{Supervising K-Means Nystr\"{o}m}
 For K-Means Nystr\"{o}m on large datasets, it is standard to train the k-means with a subset of the data sampled uniformly at random. We select this subset of samples using the negative margin, before training the final classifier. Then K-Means is applied to the chosen subset of samples in step 5 of Algorithm \ref{alg:simple}. This procedure significantly alters the way supervision is applied, which for other Nystr\"{o}m variants, is to select the support vectors directly using their approximate margin values. 
 
 In K-Means Nystr\"{o}m, the data is projected to a subspace spanned by the cluster centroids, which may be different from the subspace spanned by the samples themselves. Therefore, this Nystr\"{o}m variant is not as suitable for the application of support vector selection. Nonetheless, we provide the results for supervised K-Means Nystr\"{o}m here for comparison.
 
 \subsection{Experimental Setup}
 Similar to our experiments with Standard and Ensemble Nystr\"{o}m, we set $k=n$ to exploit the full space spanned by the representative samples, in this case, the cluster centroids.
 
 It was suggested in \cite{zhang2008improved} to perform K-Means with $20000$ randomly chosen samples on large datasets. Hence, we do so on Shuttle, COD-RNA, MNIST and Buzz data. For the implementation of K-Means Nystr\"{o}m with support vector selection (\emph{KNys+}), we select $5000$ samples from HAR, $10000$ samples from Letter, and $20000$ samples from all the other datasets as inputs to k-means. Therefore, the inputs of k-means end up being supervised, instead of the inputs of KPCA. We set $n_0=n_f$, meaning the same numbers of cluster centroids were used to train KPCA in the first and second stages of Algorithm \ref{alg:simple}.
 
 \subsection{Results}
 \begin{figure*}[t]
\begin{minipage}[b]{0.32\linewidth}
  \centering
  \centerline{\includegraphics[width=5.0cm]{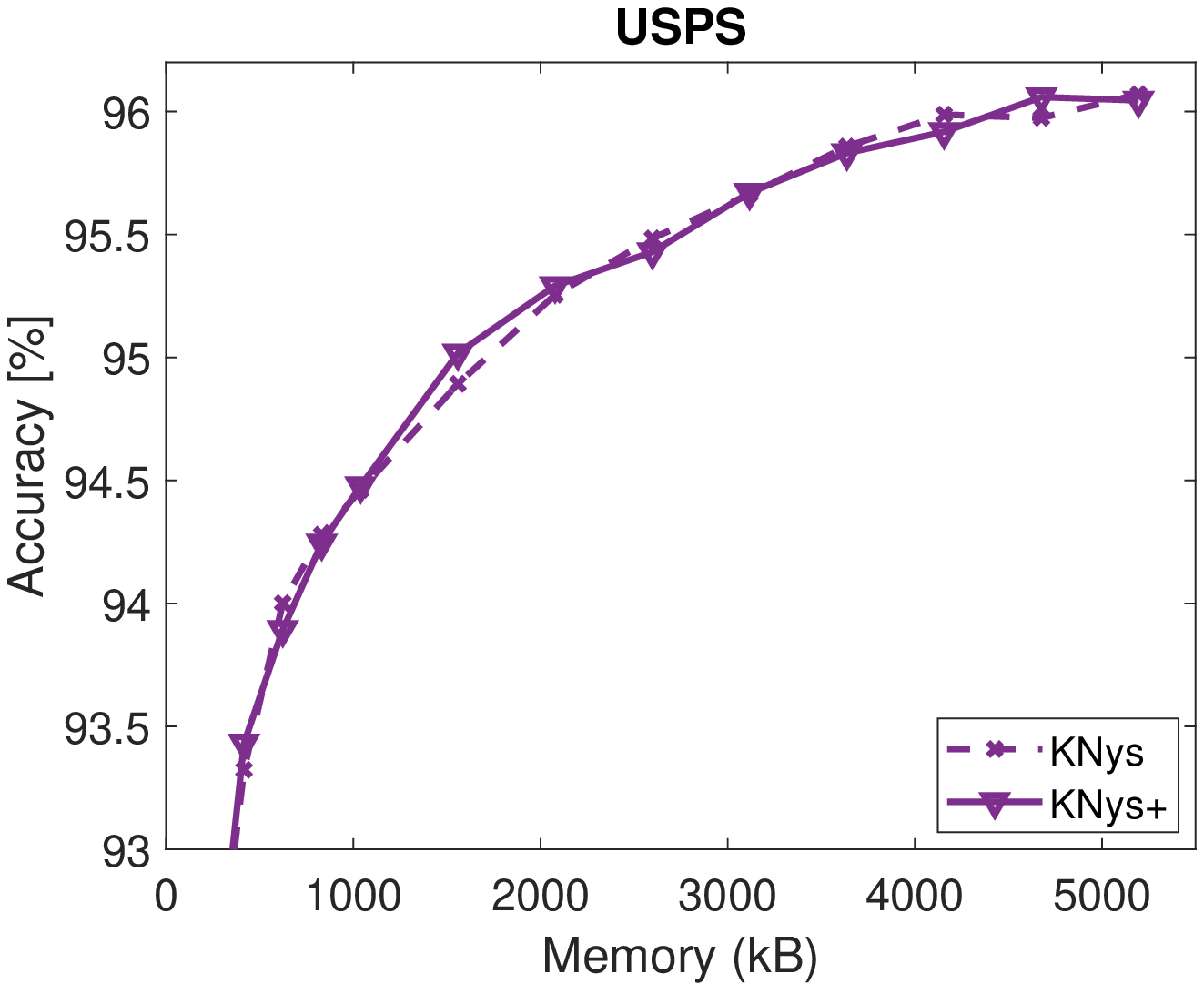}}
\end{minipage}
\hfill
\begin{minipage}[b]{0.32\linewidth}
  \centering
  \centerline{\includegraphics[width=5.0cm]{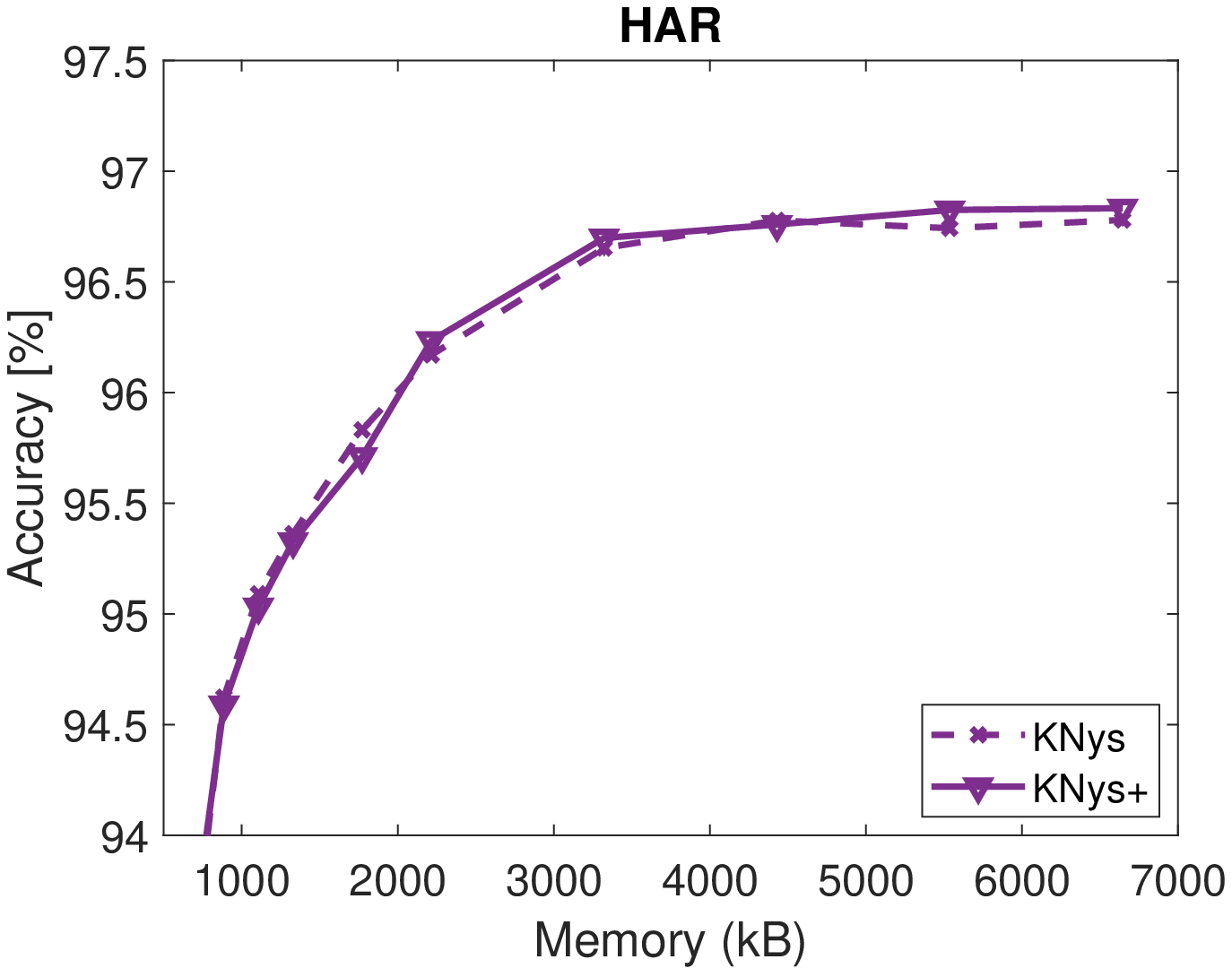}}
\end{minipage}
\hfill
\begin{minipage}[b]{0.32\linewidth}
  \centering
  \centerline{\includegraphics[width=5.0cm]{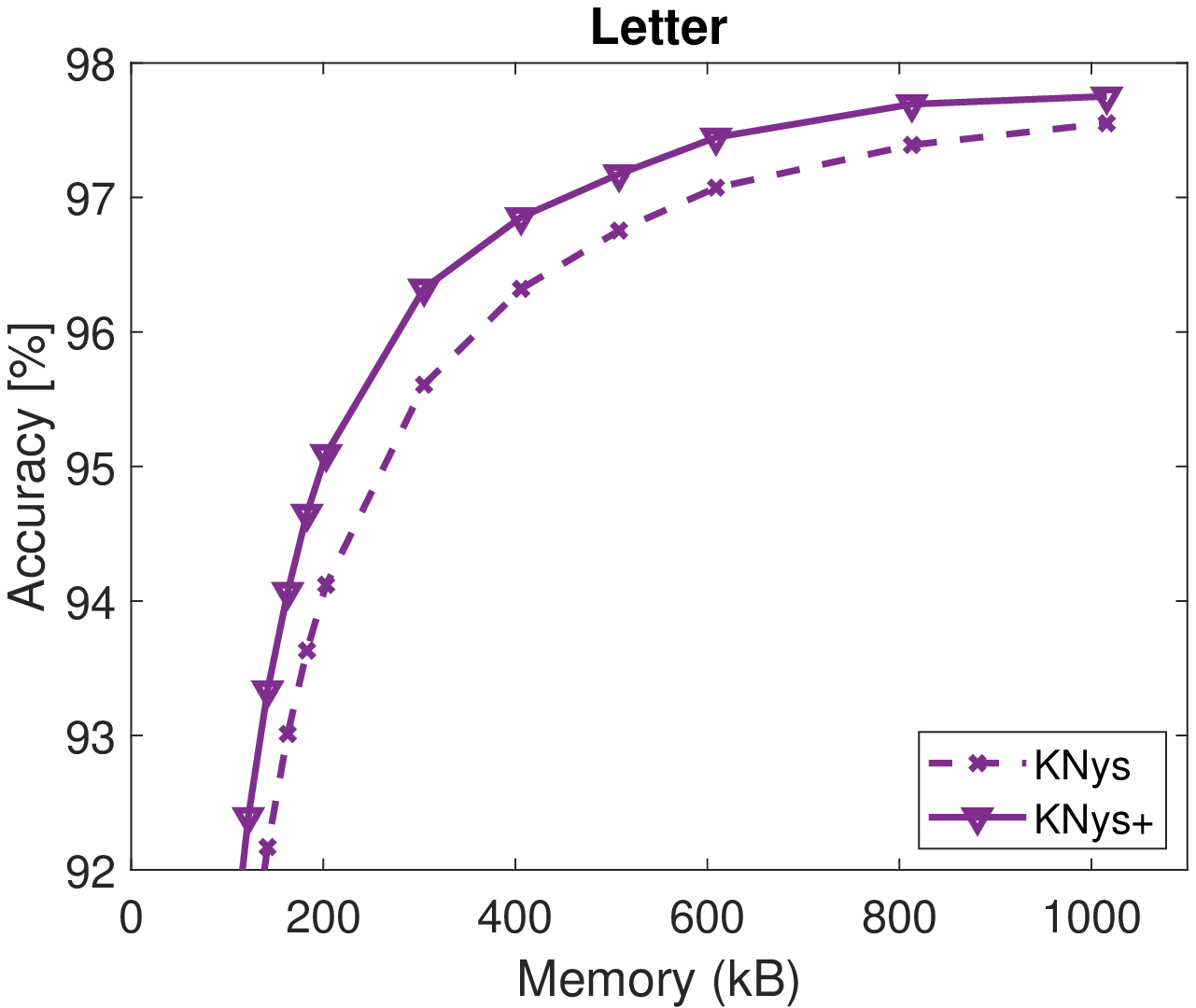}}
\end{minipage}
\begin{minipage}[b]{0.32\linewidth}
  \centering
  \centerline{\includegraphics[width=5.0cm]{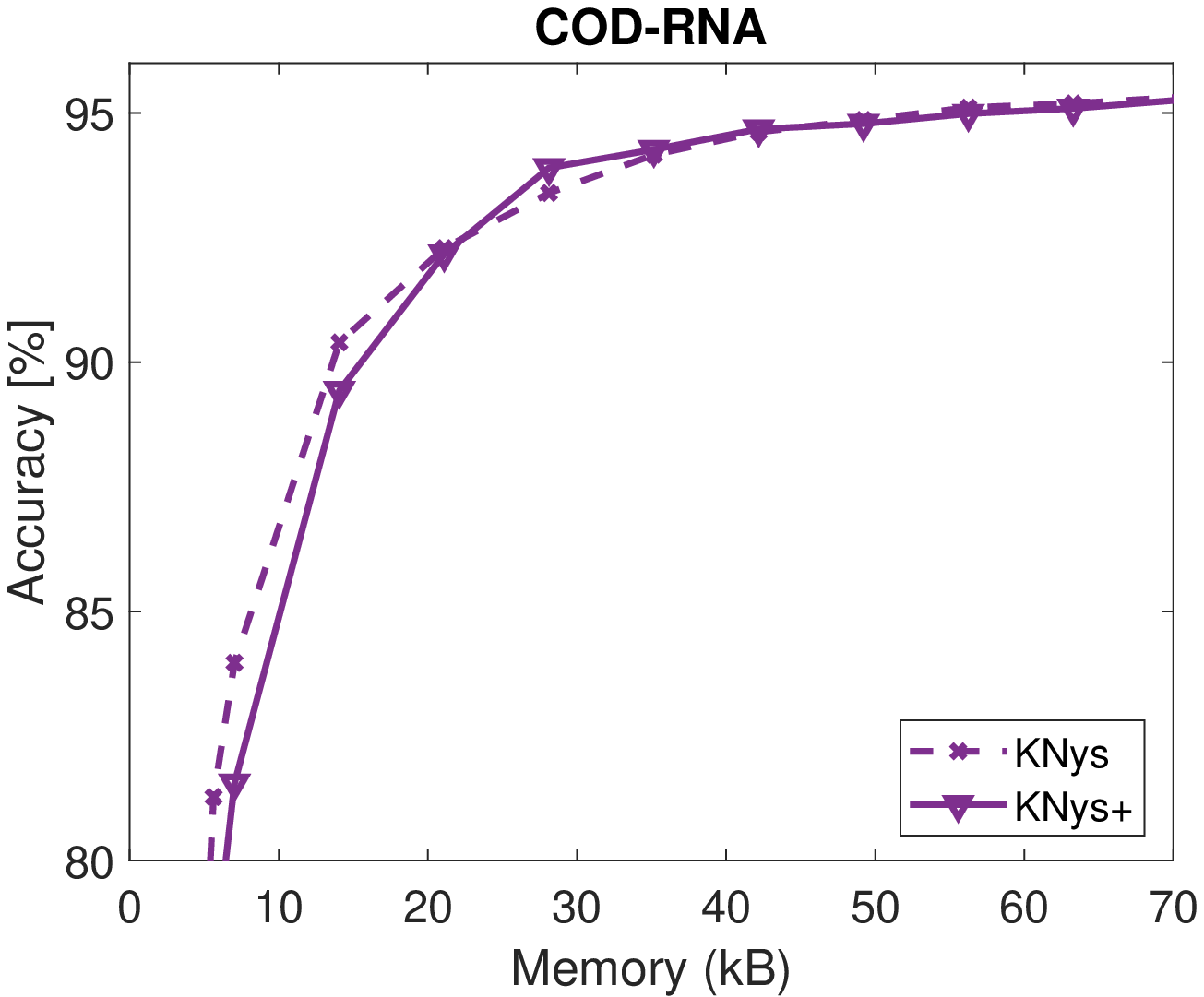}}
\end{minipage}
\hfill
\begin{minipage}[b]{0.32\linewidth}
  \centering
  \centerline{\includegraphics[width=5.0cm]{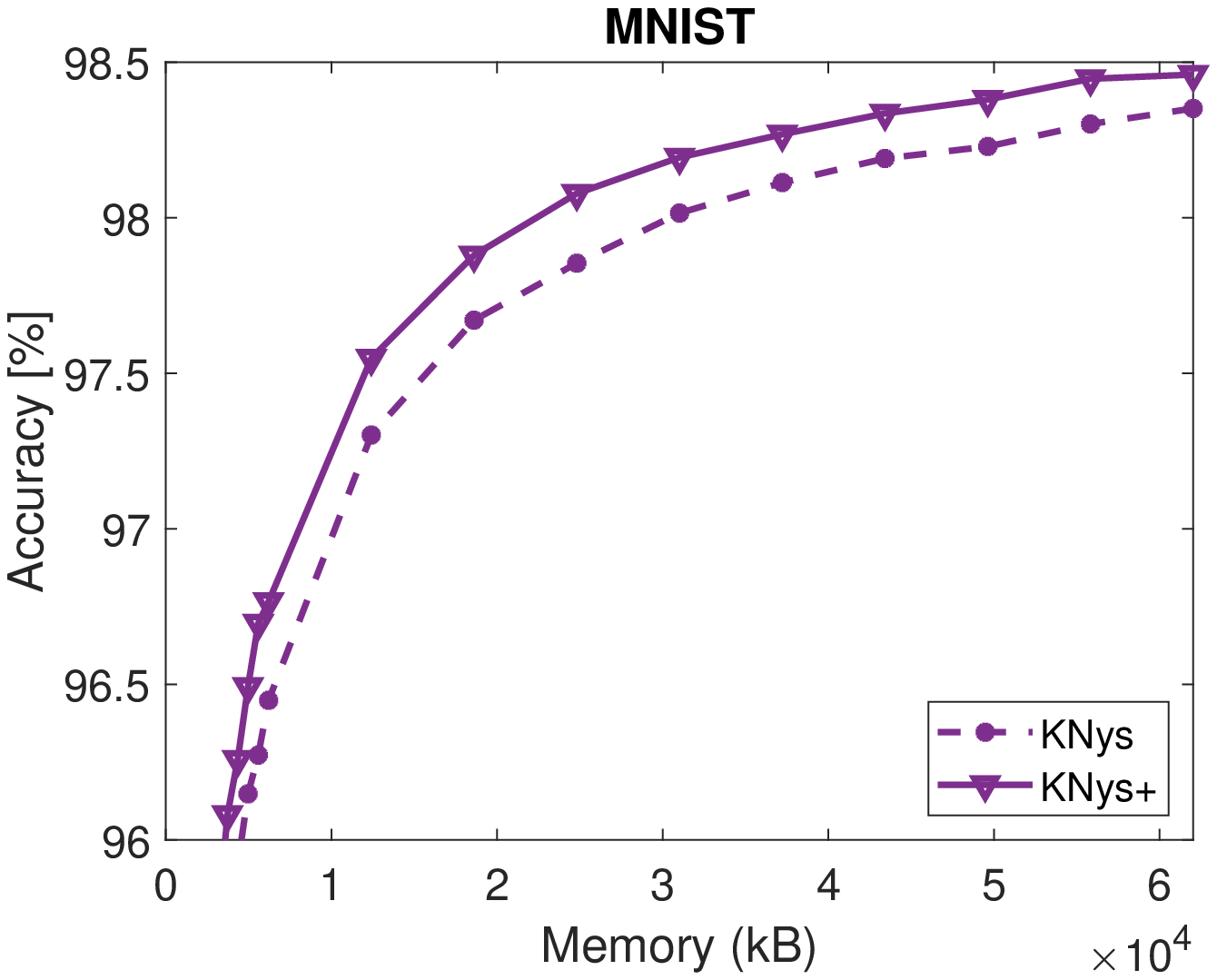}}
\end{minipage}
\hfill
\begin{minipage}[b]{0.32\linewidth}
  \centering
  \centerline{\includegraphics[width=5.0cm]{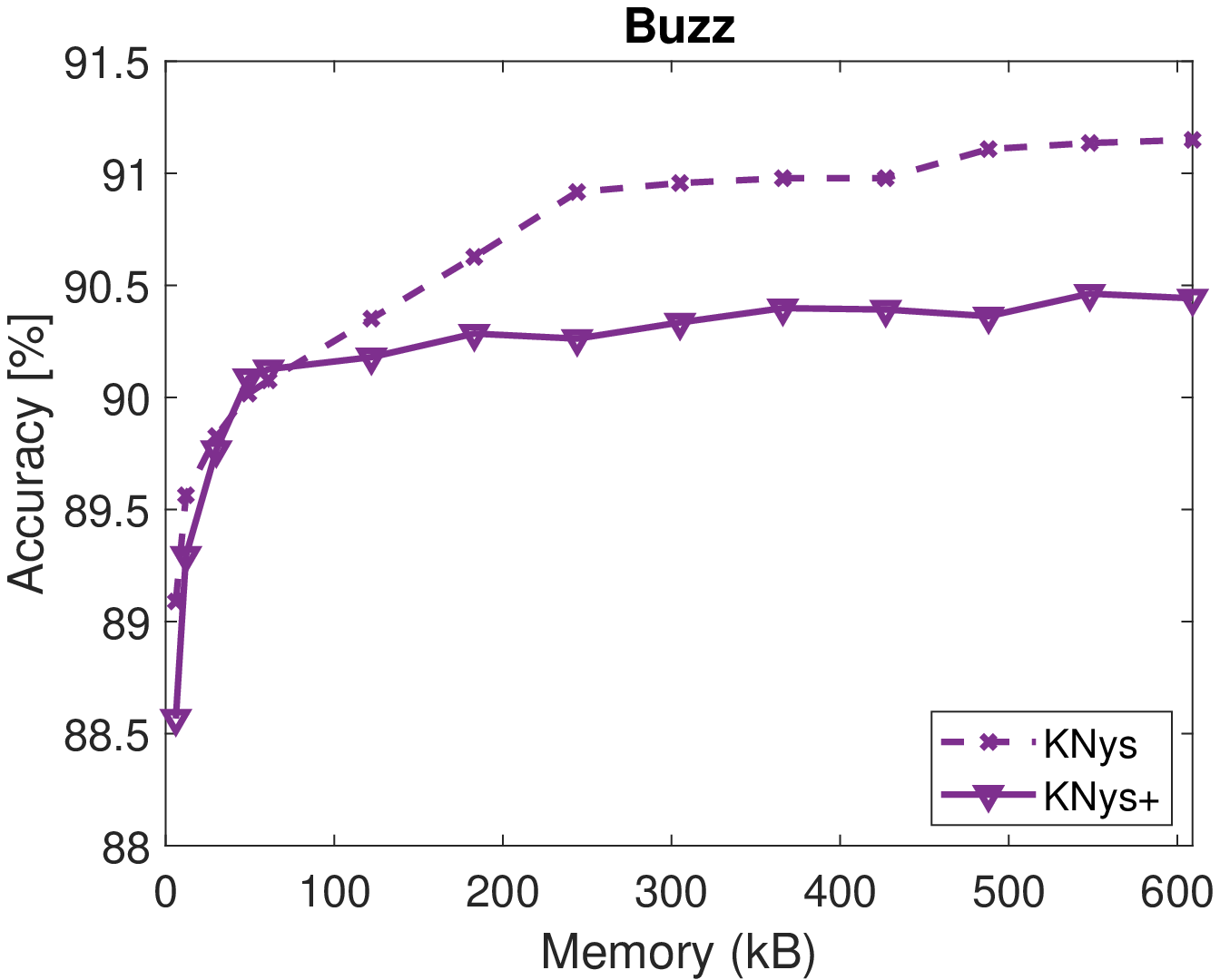}}
\end{minipage}
\caption{Size of the classifier model vs. prediction accuracy, using K-Means Nystr\"{o}m and K-Means Nystr\"{o}m with support vector selection.}
\label{fig:K-Means}
\vspace{-5pt}
\end{figure*}
 
 As demonstrated in Figure \ref{fig:K-Means}, obtaining k-means centroids from a chosen subset of samples only leads to improvement on 2 of the 6 datasets; Letter and MNIST. This is despite the fact that the chosen support vectors work very well on all 6 datasets, when they are not clustered by k-means. A possible explanation for this is that centroids retain high negative margins on Letter and MNIST, whereas they lose this characteristic in the kernel feature space on the other datasets. Therefore, clustering the support vectors after selection can be less effective, which prompts us to perform selection after clustering, that is, to ensure that the chosen cluster centers have high negative margins.
 
 In the next section, we consider supervised selection being applied to the outputs instead of the inputs of the k-means. We find that this approach works well, and can be a good way to deal with data redundancy by ensuring that successive support vectors add more information to the model.
 
 \section{Support Centroid Selection for K-Means Nystr\"{o}m} \label{app:support_centroid}
 \setcounter{table}{1}
 \begin{table}[t]
  \caption{Summary of the datasets used in additional experiments, $\gamma$ is the RBF kernel parameter.}
  \vspace{2mm}
  \centering
  \begin{tabularx}{\linewidth}{l Y Y Y Y Y}
  \toprule
  Dataset & \# Features & \# Training & \# Testing & \# Classes & $\gamma$ \\
  \midrule
  SVHN \cite{netzer2011reading} & $3072$ & $73257$ & $26032$ & $10$ & $0.001$ \\
   IJCNN \cite{chang2011libsvm} & $22$ & $120000$ & $21691$ & $2$ & $0.1$ \\
   CovType \cite{blackard1998comparative} & $54$ & $500000$ & $81012$ & $7$ & $1.0$ \\ 
   \bottomrule
  \end{tabularx}
  \label{tab:Datasets2}
\end{table}

 \begin{figure*}[t]
\begin{minipage}[b]{0.32\linewidth}
  \centering
  \centerline{\includegraphics[width=5.0cm]{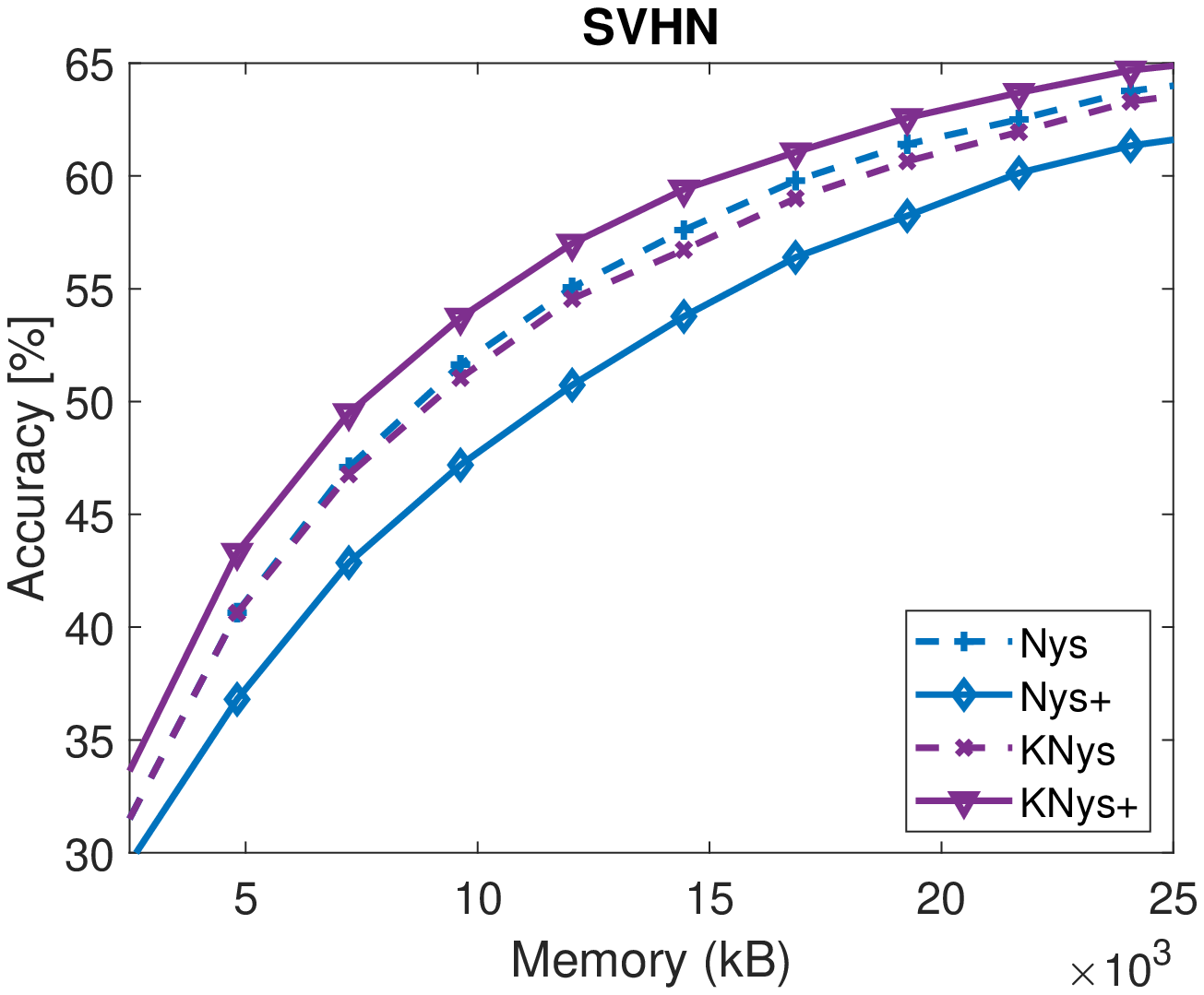}}
\end{minipage}
\hfill
\begin{minipage}[b]{0.32\linewidth}
  \centering
  \centerline{\includegraphics[width=5.0cm]{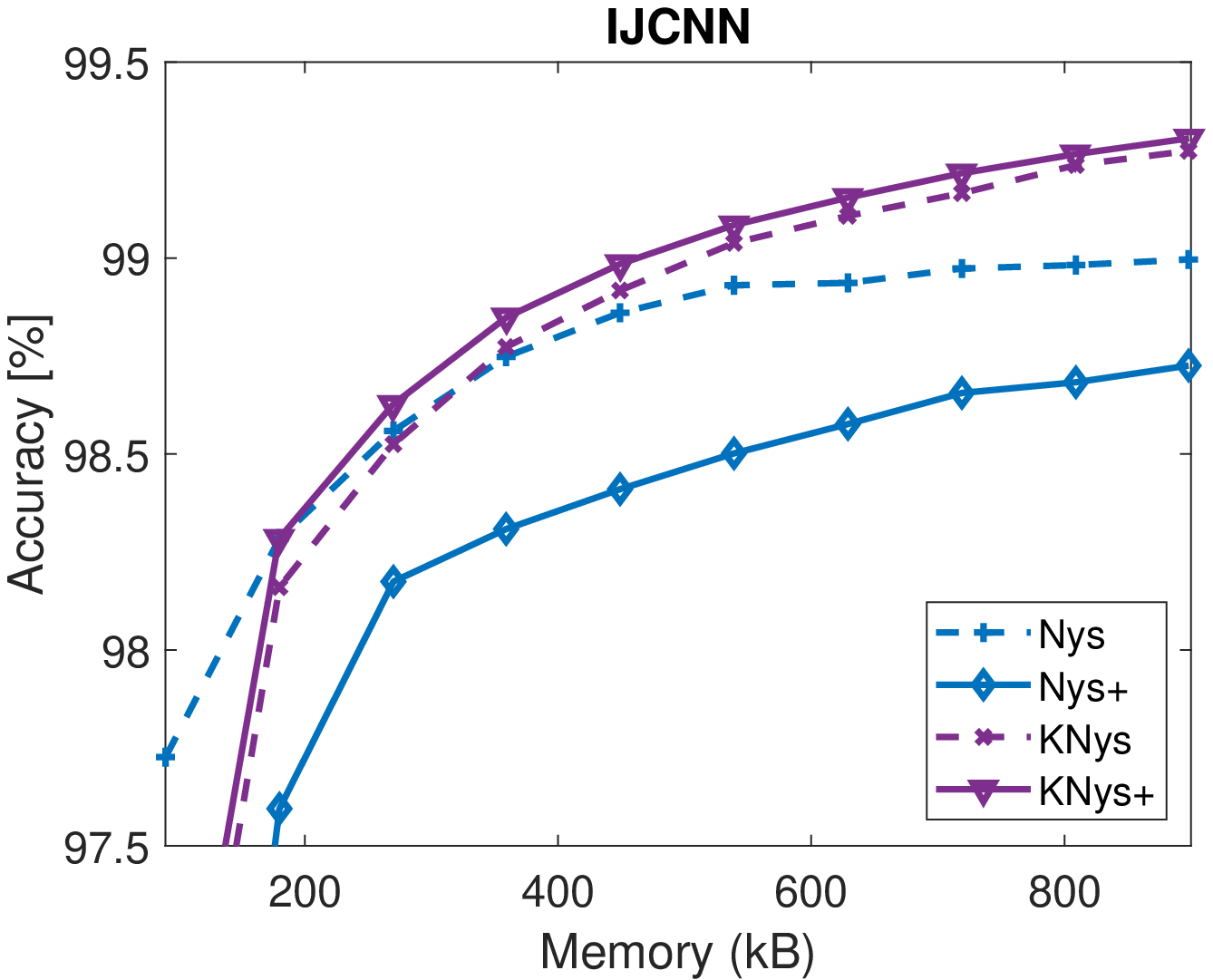}}
\end{minipage}
\hfill
\begin{minipage}[b]{0.32\linewidth}
  \centering
  \centerline{\includegraphics[width=5.0cm]{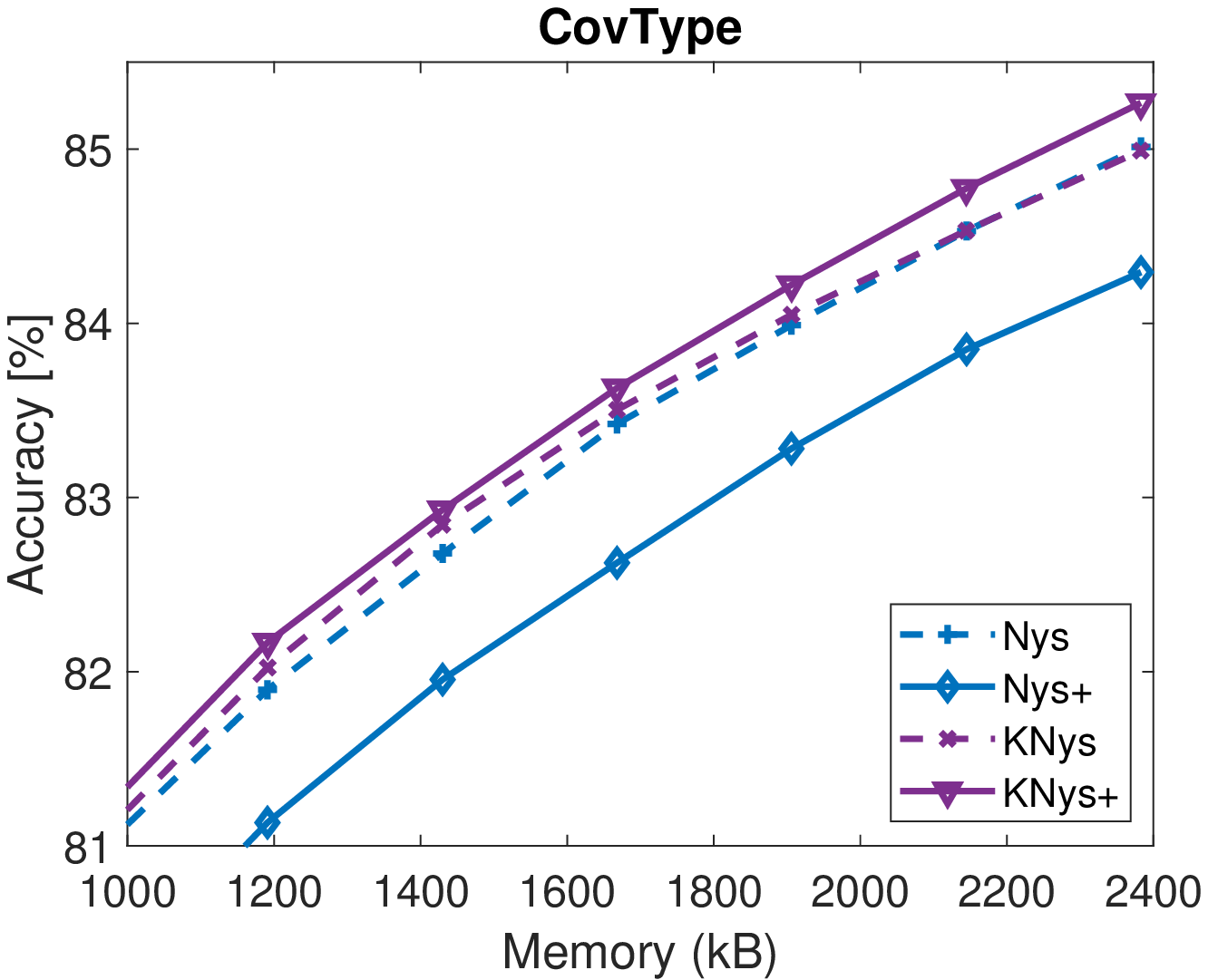}}
\end{minipage}
\caption{Size of the classifier model vs. prediction accuracy, using Standard Nystr\"{o}m, Nystr\"{o}m with support vector selection, K-Means Nystr\"{o}m, and K-Means Nystr\"{o}m with support centroid selection.}
\label{fig:K-Means+}
\vspace{-5pt}
\end{figure*}

 Some datasets may contain a large number of similar samples, which could be problematic for our support vector selection method. For instance, if a dataset contains duplicate entries, they will be given the same margin values, which can lead to the same projection directions being applied multiple times. While standard Nystr\"{o}m techniques can alleviate data redundancy to a large extent by sampling uniformly at random, our support vector selection scheme needs to be modified to work well on datasets with redundant entries.

 In this section, we modify the K-Means Nystr\"{o}m to perform support vector selection among the cluster centroids, which is a possible way to deal with redundancy. We then show the results on 3 datasets, where this method leads to improvement, while support vector selection alone does not.

\subsection{Supervised Centroid Selection}
 To perform centroid selection using the negative margin, the k-means centroids need to be assigned labels. This is done by a voting scheme. The label of each cluster centroid is determined by the majority vote of the members of its cluster. Afterwards, negative margin based support vector selection is applied to the cluster centroids, instead of the original training samples.
 
 \subsection{Experimental Setup}
 We perform the centroid selection experiments on 3 different datasets, which are summarized in Table \ref{tab:Datasets2}. $20000$ samples are used for the k-means clustering on SVHN and IJCNN, $50000$ samples are used for the k-means clustering on CovType. For K-Means Nystr\"{o}m with support centroid selection (\emph{KNys+}), we begin with $3n_f$ clusters, then $n_f$ of the centroids are selected with the negative margin criterion. To approximate the margins in the first step of Algorithm \ref{alg:simple}, we use $n_0$ randomly chosen centroids. The default setting $n_0=n_f$ is used for our experiments in this section.
 
 \subsection{Results}
Figure \ref{fig:K-Means+} compares the accuracies of Standard Nystr\"{o}m (\emph{Nys}),  Nystr\"{o}m with support vector selection (\emph{Nys+}), K-Means Nystr\"{o}m (\emph{KNys}), and K-Means Nystr\"{o}m with support centroid selection (\emph{KNys+}). The specific observations on the datasets are as follows.

\textbf{SVHN:} \emph{KNys+} performs the best and outperforms \emph{Nys} and \emph{KNys} by more than $1\%$ above $5000$ kB model size.

\textbf{IJCNN:} \emph{KNys+} performs the best and outperforms \emph{Nys} and \emph{KNys} above $200$ kB model size by more than $0.05\%$, which is less significant. However, we note that the baseline performances on this dataset are very high, being above $99\%$ for the most part.

\textbf{CovType:} \emph{KNys+} performs the best and outperforms \emph{Nys} and \emph{KNys} by $0.1\%$ to $0.27\%$ above $1000$ kB model size. The improvement increases as the model size increases for the range of sizes displayed.

Support vector selection originally fails to improve the predictive accuracy of Standard Nystr\"{o}m on these datasets. However, thanks to the reduction in sample redundancy via k-means clustering, K-Means Nystr\"om with support centroid selection outperforms both Standard and K-Means Nystr\"{o}m on all three datasets for a wide range of complexities. This is especially significant on SVHN, since K-Means Nystr\"{o}m normally performs worse than Standard Nystr\"{o}m on this data. 

Reducing data redundancy via k-means comes at a computational cost, which can be significant compared to the training time of Standard Nystr\"{o}m. Future work will incorporate redundancy reduction into the selection procedure, which might be done at a low computational cost. 

\end{appendix}

\end{document}